\documentclass{article}
\usepackage[preprint,nonatbib]{neurips_2026}
\usepackage[utf8]{inputenc}
\usepackage[T1]{fontenc}
\usepackage{hyperref}
\usepackage{url}
\usepackage{booktabs}
\usepackage{multirow}
\usepackage{amsfonts}
\usepackage{amsmath}
\usepackage{nicefrac}
\usepackage{microtype}
\usepackage{xcolor}
\usepackage{graphicx}
\usepackage{array}
\usepackage{subcaption}
\usepackage{algorithm}
\usepackage{algpseudocode}
\usepackage{amsmath}
\usepackage{float}
\usepackage{nicefrac}
\usepackage{xspace}
\usepackage{subcaption}

\title{Interpretable Adaptive Sampling for LLM Test-Time Scaling}

\author{
Mobina Kashaniyan, Ali Jannesari\\
Iowa State University\\
Ames, IA, USA\\
\texttt{\{mobina,jannesar\}@iastate.edu}
}
\begin{document}
\maketitle
\begin{abstract}
Test-time scaling improves LLM reasoning by generating and aggregating multiple candidate answers, yet many pipelines use fixed per-query budgets that spend the same compute on easy and difficult prompts. These fixed budgets are also difficult to inspect because they do not explain why a given prompt receives a particular number of samples. We propose \textit{adaptive} test-time scaling with a lightweight fuzzy controller that maps interpretable signals, including estimated prompt complexity and model confidence, to a per-query sampling budget. The controller assigns fewer samples to easier or more confident prompts and more samples to harder or less certain prompts, making inference-time compute inspectable rather than fixed or opaque. We evaluate under a fair-alignment protocol with matched decoding settings and controlled answer selection, and compare against best-of-$N$, compute-aware scaling, and self-certainty-based baselines on question-answering and mathematical reasoning tasks. Across models and datasets, adaptive fuzzy control improves over several standard baselines and remains close to a selector-matched full-budget control while reducing the average number of samples. These findings suggest that interpretable adaptive sampling is a practical direction for more efficient test-time reasoning in large language models.
\end{abstract}

\section{Introduction}

Large language models (LLMs) achieve strong results on many reasoning and question-answering tasks, but their inference behavior is still difficult to inspect. In practice, models are often used as black boxes: we observe the prompt and the final answer, but we usually do not know why a query should receive more reasoning effort, more samples, or a different decoding configuration than another query \cite{singhc,bboxllm}. This matters for interpretability, debugging, safety review, and deployment under limited compute budgets. A complementary line of work improves LLM performance by spending more computation at inference time. Test-time scaling generates multiple candidate answers, aggregates them with voting or reranking, or adjusts decoding parameters such as temperature and output length. These methods can improve accuracy on difficult reasoning tasks without changing model weights \cite{wang2023selfconsistency,snell2024scaling}. However, many test-time scaling pipelines still rely on a fixed sampling budget. Every prompt receives the same number of candidate answers, even though prompts vary widely in difficulty and uncertainty. This can waste computation on easy instances while still underserving hard ones. A second challenge is that compute allocation is often opaque. A fixed budget is easy to implement, but it does not explain why a specific prompt received that amount of computation. Learned rerankers or policy models may improve performance, but their allocation decisions can be difficult to audit. For adaptive inference to be useful in practice, the system should not only decide how much compute to spend, but also make that decision understandable.

This paper studies adaptive test-time scaling with an explicit and interpretable budget controller. We ask the following question:
\begin{quote}
\emph{Can an interpretable controller allocate test-time samples per prompt so that an LLM preserves most of the accuracy of full-budget sampling while using fewer samples on average?}
\end{quote}

This question has two parts. First, the controller must identify which prompts can safely use fewer samples and which prompts should remain close to the full budget. Second, the evaluation must separate the effect of the budget policy from the effect of the answer selector. To address this, we use a hierarchical fuzzy controller that maps human-readable signals, such as estimated prompt complexity, confidence, entropy, prompt type, expected answer length, and lightweight historical performance, to a per-prompt sampling budget. The controller outputs a continuous scale value, which is converted into an integer sample count. We then evaluate it under a fair-alignment protocol where decoding settings are matched, the answer selector is controlled, and the adaptive method changes only the number of generated candidates.

Our contributions are as follows:
\begin{itemize}
  \item \textbf{Adaptive test-time scaling with interpretable control.} We introduce a hierarchical fuzzy controller that allocates computation per prompt based on simple and understandable signals, providing a transparent alternative to fixed or opaque scaling strategies.
\item \textbf{Unified pipeline for scaling, generation, and selection.} We design an end-to-end framework that integrates adaptive sampling with self-certainty scoring and Borda aggregation, allowing better use of multiple candidate answers.
     \item \textbf{Improved accuracy--compute tradeoff.} We show that our method achieves competitive or improved accuracy compared to fixed-budget baselines while using fewer samples on average, focusing computation where it is most useful.

\item \textbf{Explainable and interpretable decision-making.} The fuzzy controller provides a clear mapping from input signals to scaling decisions, allowing the reasoning process to be inspected and understood, unlike typical black-box approaches.

\end{itemize}

Together, these results support a simple claim: test-time scaling should not only add more computation, but should allocate it in a way that is effective, controlled, and inspectable.
\section{Related Work}
\label{sec:related-work}
\paragraph{Test-time scaling for LLM reasoning.}
Recent work has shown that LLM reasoning can improve when additional computation is spent at inference time. Self-consistency samples multiple reasoning paths and selects the most common final answer \cite{wang2023selfconsistency}. Snell et al.~\cite{ttcompute} show that test-time compute can sometimes be more effective than increasing model size when the compute budget is allocated well. Liu et al.~\cite{1bllm} further show that the benefit of test-time scaling depends on the interaction among the policy model, reward model, and problem difficulty. A recent survey by Zhang et al.~\cite{ttssurvey} organizes this growing literature around what is scaled, how scaling is performed, where it is applied, and how it should be evaluated. \paragraph{Structured and adaptive inference.}
Several studies propose more structured forms of inference-time reasoning. Teng et al.~\cite{atomofthought} decompose reasoning into atomic steps to reduce redundant dependence on long reasoning histories. Wang et al.~\cite{mctsjudge} use Monte Carlo Tree Search for LLM-as-a-judge code evaluation. Huang et al.~\cite{selfcalib} use self-calibrated confidence to make test-time scaling more efficient. Other work studies when additional reasoning can hurt. Liu et al.~\cite{rethinkingprompt} show that complex prompting strategies may lose their advantage as sampling increases, while Yang et al.~\cite{optimaltts} caution that excessive reasoning can cause overthinking. These findings motivate adaptive methods that decide when additional computation is useful rather than applying the same budget to every prompt. \paragraph{Scaling beyond text reasoning.}
Test-time scaling has also been studied in multimodal models, code systems, software agents, GUI agents, and edge deployment. Work on multimodal and world-foundation models studies whether additional inference compute improves visual and multimodal reasoning \cite{expandperfboundaries,ttsfoundationmodels}. Agentic systems such as Trae Agent, GTA1, and General AgentBench examine scaling in settings where models must plan, act, and interact with tools or interfaces \cite{traellm,guitts,benchmarktts}. Systems-oriented work such as Kinetics and FastTTS highlights practical constraints including latency, memory access, and hardware efficiency \cite{kinetics,fasttts}. These studies show that test-time scaling is useful but also costly, making adaptive compute allocation important. \paragraph{Interpretability and adaptive control.}
Our work is closest to adaptive compute allocation, but differs in its emphasis on interpretability and fair comparison. Rather than learning a black-box budget policy, we use a fuzzy controller whose inputs, rules, and outputs can be inspected for each prompt. This connects to interpretability work arguing that LLM behavior should be made more
understandable, while also warning that generated explanations may be unfaithful to the
model's actual decision process \cite{notsay}. We treat test-time scaling not only as a way to improve performance, but also as an interpretable decision process for adaptive inference.
\section{Proposed Method}
\label{sec:method}

\begin{figure}[h]
    \centering
\includegraphics[width=1\linewidth]{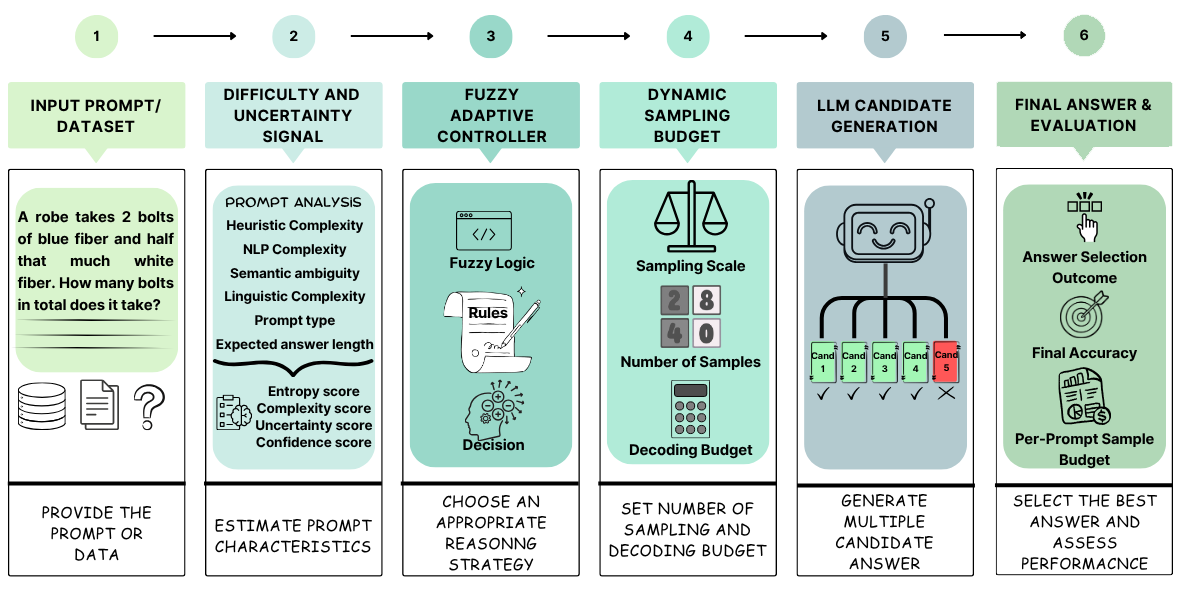}
\caption{Adaptive test-time scaling pipeline. The system estimates prompt difficulty and uncertainty, assigns a per-prompt sampling budget with a fuzzy controller, generates candidates, and selects the final answer.}
    \label{fig:pipeline}
\end{figure}

Figure~\ref{fig:pipeline} shows the proposed adaptive test-time scaling pipeline. Given a prompt $x$, the system extracts difficulty and uncertainty signals, sends them to a hierarchical fuzzy controller, and converts the controller output into an integer sampling budget $N(x)$. The LLM then generates $N(x)$ candidate answers, and a fixed selector chooses the final response. The pipeline has six stages: input prompt, difficulty and uncertainty estimation, fuzzy adaptive control, dynamic budget selection, candidate generation, and final answer selection.
\subsection{Signal Extraction}
All controller inputs are normalized to $[0,1]$. We use a mix of prompt-side signals, which are available before generation, and model-side uncertainty signals, which are estimated from an initial short draft. The first prompt-side signal is a surface-level complexity score. Let $n_w$ be the number of words in the prompt. We compute a normalized length term
\begin{equation}
\ell = \mathrm{clamp}\left(\frac{n_w}{60}\right),
\end{equation}
and add small bonuses for punctuation, digits, uppercase-heavy text, and multiple sentences. This score, denoted by $h$, captures the intuition that longer and more structured prompts tend to require more computation. We also compute a lightweight NLP-based complexity score. This score combines normalized length, vocabulary richness, sentence count, math-symbol density, semantic ambiguity, linguistic complexity, and reasoning depth:
\begin{equation}
 c_{\mathrm{NLP}} = \min\Bigl(1,0.15\tilde\ell+0.10\tilde r+0.10\tilde s+0.15\tilde m+0.20a_{\mathrm{sem}}+0.15a_{\mathrm{ling}}+0.15a_{\mathrm{rea}}\Bigr).
\end{equation}
The final complexity score is
\begin{equation}
 c = \mathrm{clamp}\bigl(0.6c_{\mathrm{NLP}}+0.4h\bigr).
\end{equation}
We use this weighted combination because surface features are stable and inexpensive, while the NLP features provide a richer description of ambiguity and reasoning demand. The controller also uses prompt type $\tau$ and expected answer length $\lambda$. The prompt type signal separates short factual questions from open-ended or reasoning-heavy prompts using keyword and structural cues. The expected answer length signal estimates whether a prompt likely requires a short answer, a derivation, or a longer explanation. For model-side uncertainty, the system generates a short draft answer and computes a confidence score from token probabilities:
\begin{equation}
\gamma = \mathrm{clamp}\left(\exp\left(\frac{1}{L}\sum_{i=1}^{L}\log p_i\right)\right),
\end{equation}
where $L$ is the draft length and $p_i$ is the probability assigned to token $i$. A higher $\gamma$ indicates that the model assigned higher probability to its generated tokens. We also compute normalized entropy $\eta$, where larger values indicate a more diffuse token distribution. Finally, the history cache stores a slow-moving estimate $\pi$ of prior performance on similar prompts, indexed by coarse bins of complexity and prompt type. The full feature vector is
\begin{equation}
\phi=(c,\gamma,\tau,\lambda,\eta,\pi,a_{\mathrm{sem}},a_{\mathrm{ling}},a_{\mathrm{rea}}).
\end{equation}

\subsection{Fuzzy Adaptive Control}

\begin{figure}[h]
    \centering

    \begin{subfigure}[h]{0.48\linewidth}
        \centering
        \includegraphics[width=\linewidth]{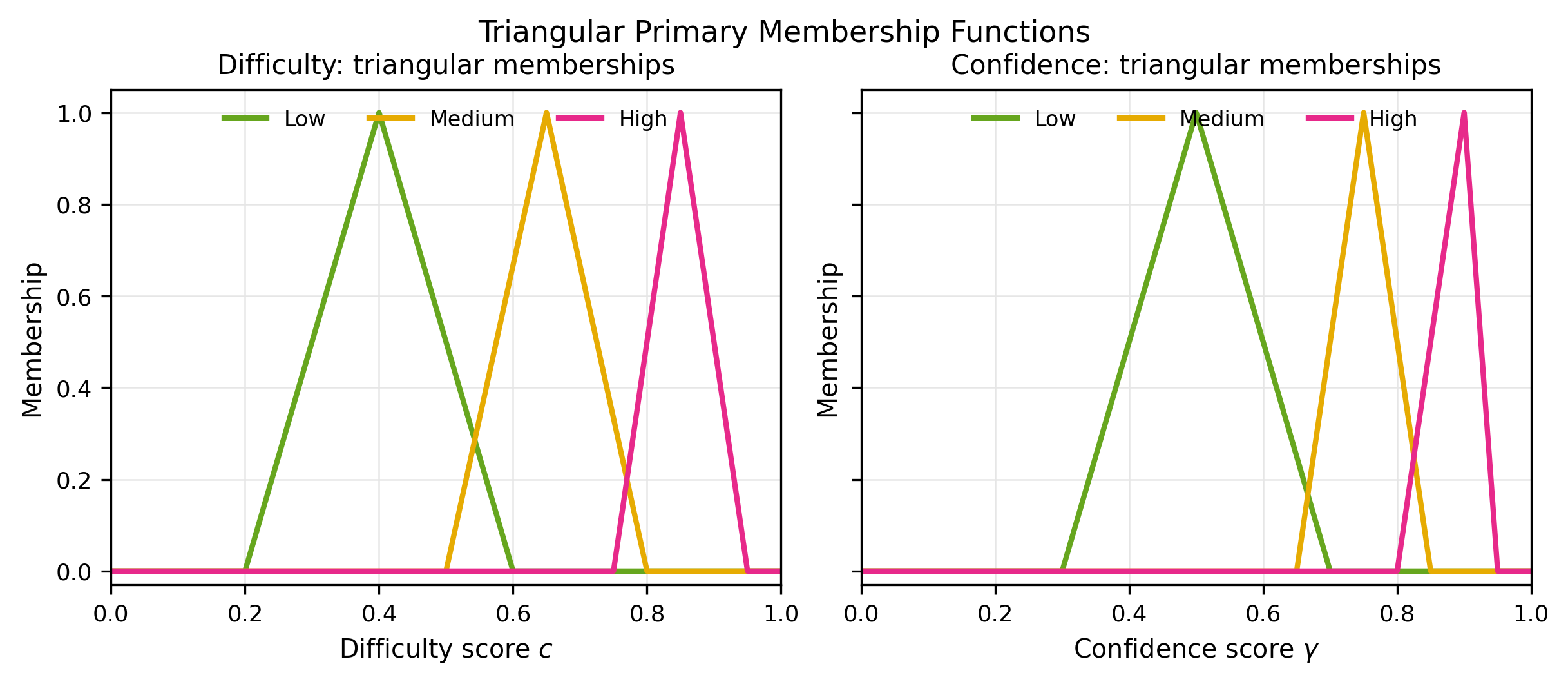}
        \caption{Triangular memberships for low, medium, and high regions.}
        \label{fig:fuzzy-triangular}
    \end{subfigure}
    \hfill
    \begin{subfigure}[h]{0.48\linewidth}
        \centering
        \includegraphics[width=\linewidth]{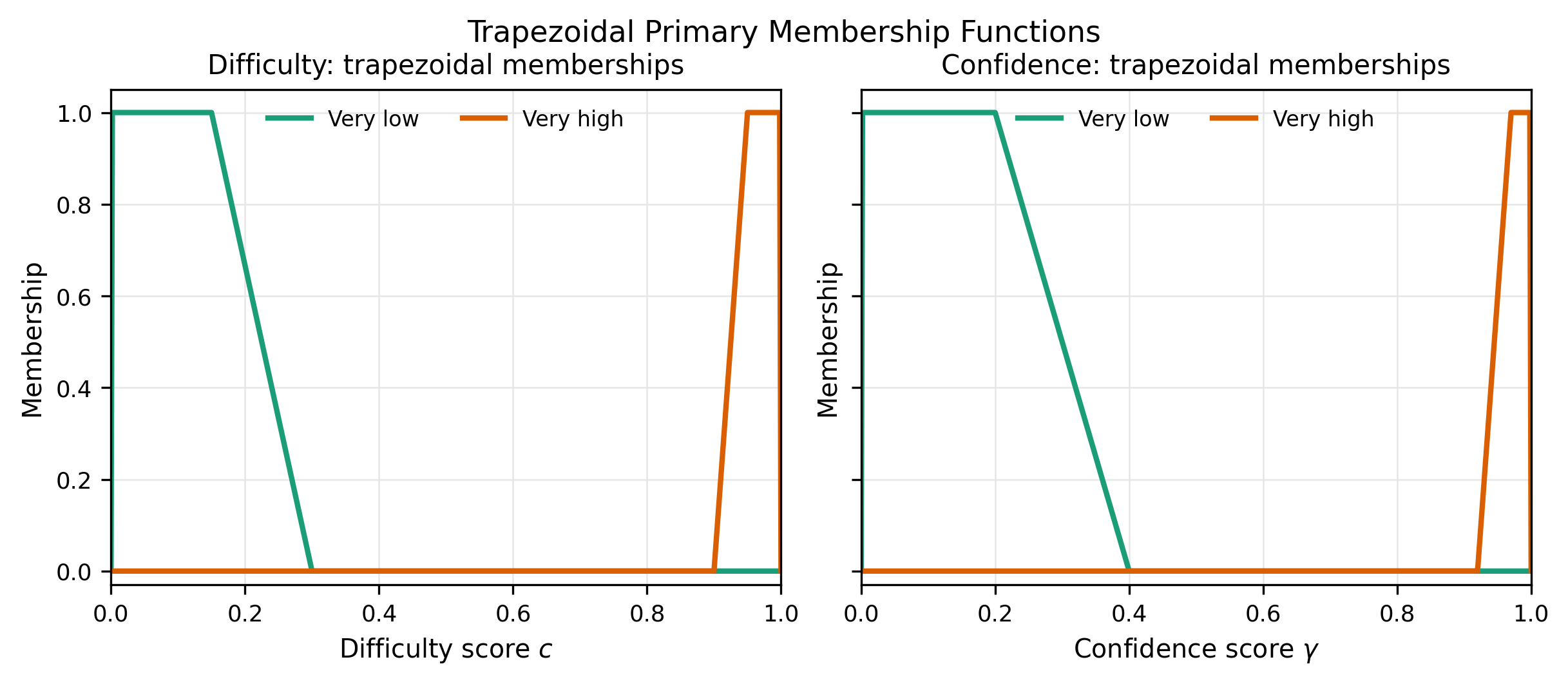}
        \caption{Trapezoidal memberships for very-low and very-high regions.}
        \label{fig:fuzzy-trapezoidal}
    \end{subfigure}

    \caption{Primary fuzzy membership functions used by the controller for difficulty and confidence. The controller combines triangular and trapezoidal membership functions in the same fuzzy rule system: triangular functions represent gradual low, medium, and high regions, while trapezoidal functions capture very-low and very-high extremes.}
    \label{fig:fuzzy-memberships}
\end{figure}

The fuzzy controller outputs two values: a scale $s\in[0,1]$ and an uncertainty value $u\in[0,1]$. The scale controls how much compute the prompt receives. The uncertainty value measures how wide the Type-2 fuzzy intervals are after refinement. The first stage uses the two strongest signals: complexity $c$ and confidence $\gamma$. Each signal is mapped to fuzzy labels such as low, medium, and high using triangular and trapezoidal membership functions. The rule base then activates compute labels such as very low, low, medium, high, and very high. High complexity or low confidence pushes the output toward larger budgets. Low complexity and high confidence push the output toward smaller budgets. This stage gives the controller a coarse budget decision. The second stage refines the first-stage decision. High entropy adds compute because it suggests that the model was uncertain during generation. A longer expected answer adds compute because longer answers create more opportunities for errors. Open-ended prompt type adds mild exploration. Poor historical performance on similar prompts also increases the budget. Semantic ambiguity, linguistic complexity, and reasoning depth each add a smaller refinement. These refinements are represented as Type-2 intervals. When the refinements agree, the output interval is narrow; when they conflict, the interval is wider. The final scale is obtained by defuzzification. We take a weighted average of interval centers, using the first-stage activation strengths as weights. The uncertainty $u$ is computed from the average interval width. This gives a single budget signal while preserving an interpretable measure of controller uncertainty.
Figure~\ref{fig:fuzzy-memberships} illustrates how the controller converts continuous difficulty and confidence scores into fuzzy regions. A prompt can partially belong to more than one region, such as medium and high difficulty, instead of being forced into a single discrete class. This makes the budget decision gradual: as difficulty increases or confidence decreases, the sampling budget can increase smoothly rather than changing abruptly at a fixed threshold.

\subsection{Budget Mapping and Answer Selection}
The scale is first nudged upward when the controller is uncertain:
\begin{equation}
 s'=\mathrm{clamp}(s+\alpha u), \qquad \alpha=0.2.
\end{equation}
We then map $s'$ to an integer sample budget:
\begin{equation}
N(x)=\max\left(1,\mathrm{round}\left(1+s'(N_{\max}-1)\right)\right).
\end{equation}
When a prompt is clearly hard ($c>0.6$) and $N_{\max}\ge6$, we also apply a small floor $N(x)\ge\min(6,N_{\max})$. This prevents obviously hard prompts from receiving very small budgets. After generating $N(x)$ candidates, we apply answer selector. Each candidate receives a self-certainty score based on token probabilities. Candidates are ranked by this score, final answer strings are extracted, and Borda aggregation is applied over the extracted answers. The answer with the highest total Borda score is returned \cite{borda}.

\subsection{Why Fuzzy Control Is Useful}
The reason for using fuzzy control is not that fuzzy logic is more powerful than a learned policy. A learned policy could probably fit the training traces more closely. The reason is that the budget decision in test-time scaling is naturally gradual rather than binary. A prompt is rarely only ``easy'' or only ``hard.'' It may be short but mathematically dense, or long but straightforward, or syntactically simple while still uncertain for the model. Fuzzy membership functions allow these cases to be represented as partial memberships in several difficulty and confidence regions at the same time. This makes the controller less brittle than a set of hard thresholds. This design also makes the allocation auditable. For any prompt, the system can report which signals were large, which fuzzy rules were activated, how much each refinement changed the output, and why the final sample count was selected. This is important because adaptive compute allocation can otherwise become another black-box component inside an already hard-to-interpret LLM pipeline. In our setting, interpretability means that a user can inspect the budget decision without retraining the system or interpreting a large neural policy. The controller is therefore meant to be a simple and transparent policy class for studying adaptive test-time scaling.

\section{Experimental Setup}
\label{sec:experiments}
We evaluate whether adaptive budgeting improves accuracy for a given amount of inference compute. The main evaluation uses exact-match accuracy on extracted final answers and average sample count $\bar N$ as a compute proxy. We organize the evaluation around three questions. First, does the adaptive controller improve accuracy compared with fixed-budget and compute-allocation baselines under matched decoding settings? Second, does the controller still help when the answer selector is held fixed, so that the comparison isolates the budget policy? Third, do the logged traces show behavior consistent with the intended mechanism, namely lower budgets for easy or confident prompts and higher budgets for hard or uncertain prompts? These questions are reflected in the main tables, the selector-matched control rows, and the appendix trace analyses.

\paragraph{Models and datasets.}
The main experiments use Phi-3-mini-4k-instruct and Qwen2.5-1.5B-Instruct. We evaluate on GSM8K with $1{,}319$ test questions, a $1{,}319$-item subset of MATH, and SciQ with $1{,}000$ test prompts. GSM8K \cite{gsm8k} tests grade-school arithmetic reasoning, MATH \cite{math} tests harder mathematical reasoning, and SciQ \cite{sciq} tests short factual science questions. This gives one factual QA task and two reasoning-heavy tasks.

\paragraph{Baselines.}
We compare against several fixed and compute-aware test-time scaling baselines. First, we evaluate Best-of-$N$ with $N\in\{1,3,5\}$. We also include a compute-optimal baseline inspired by Snell et al., implemented as a formula-based allocation rule. To compare against selector-based methods, we evaluate self-certainty selection at fixed $N=8$ and self-certainty+Borda at fixed $N=8$. The most important control is fixed $N=8$ with self-certainty+Borda, because it uses the same maximum sampling budget and answer-level selector as our adaptive Borda variant. This isolates the effect of adaptive budget allocation from the effect of the final selector. In additional selector-matched controls, we also compare fixed and adaptive variants using majority/plurality selection.  

\paragraph{Fair-alignment protocol.}
All main comparisons use temperature $1.0$, top-$p=0.95$, and maximum output length $256$. The adaptive method uses $N_{\max}=8$ in the main fair-alignment setting. The full system can use a short draft and can adjust decoding parameters with the scale. However, this would make comparison to fixed baselines less clean. Therefore, the main experiments use a fair-comparison mode. In this mode, decoding settings are fixed across methods, the draft pass is disabled, and the controller changes only the number of samples. This setting is stricter than the full system, but it directly answers the main scientific question: whether adaptive sample allocation helps when selection and decoding are controlled.
\section{Evaluation Results}
\label{sec:results}
\begin{figure}[h]
        \centering
        \includegraphics[width=\linewidth]{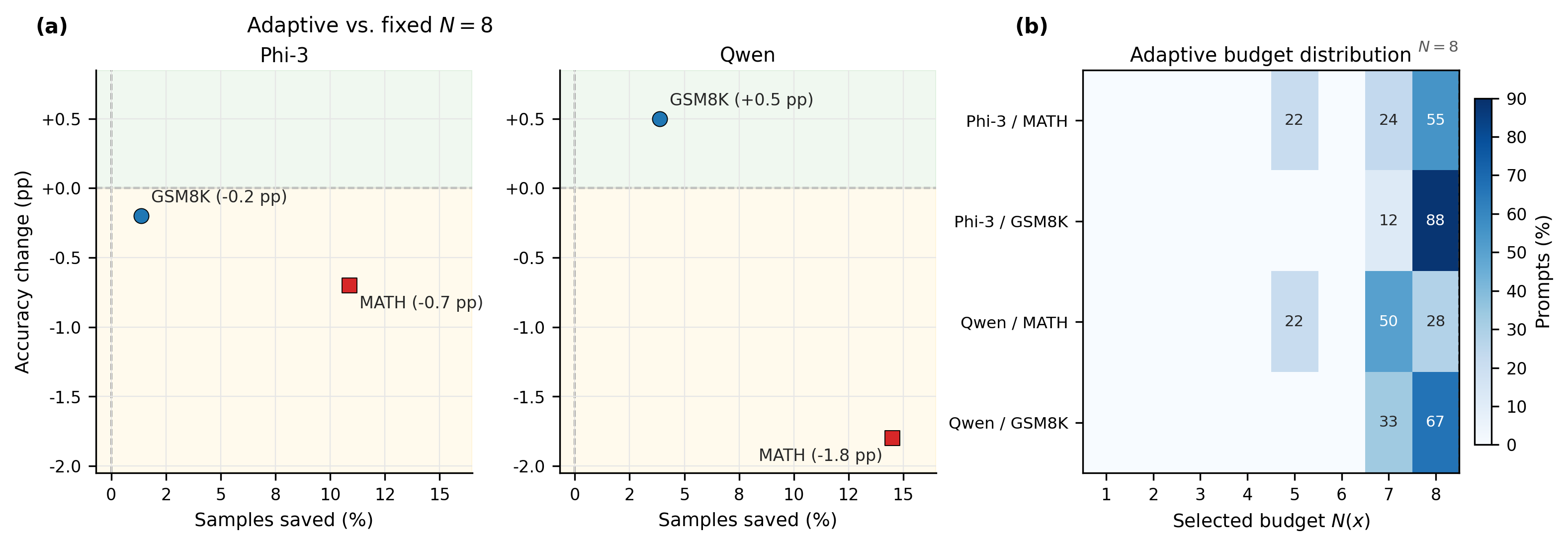}
        \label{fig:adaptive-tradeoff}
    \caption{Main evaluation results under the fair-alignment setting. \textbf{Left:} Accuracy--compute tradeoff relative to the selector-matched fixed-$N=8$ baseline. Each point compares Adaptive with fixed $N=8$ for the same model and dataset. Points to the right use fewer samples, and points above zero improve accuracy. \textbf{Right:} Distribution of selected adaptive budgets across model--dataset pairs. Darker cells indicate that a larger share of prompts received that budget. The controller uses lower budgets more often on MATH than on GSM8K, which explains the larger sample reduction on MATH.}
    \label{fig:main-results-visuals}
\end{figure}

Figure~\ref{fig:main-results-visuals} summarizes the main behavior of the adaptive controller. The left panel shows the tradeoff between sample saving and accuracy change relative to fixed $N=8$. The right panel shows how often the controller assigns each budget value. Together, these plots show that the controller does not simply use the full budget for every prompt. It saves more samples on MATH, while it behaves more conservatively on GSM8K.

\subsection{Selector-matched comparison}
Table~\ref{tab:selector-matched} gives the strongest controlled comparison. Adaptive and fixed $N=8$ use the same self-certainty+Borda selector, so the main difference is the sampling budget. This lets us test whether the adaptive controller can reduce sampling while keeping accuracy close to the full-budget baseline.
\begin{table}[h]
\centering
\caption{Selector-matched comparison between adaptive budgeting and fixed full-budget sampling. Both methods use the same self-certainty+Borda selector. Accuracy is reported with 95\% Wilson confidence intervals. $\Delta$ Acc. is Adaptive minus fixed $N=8$.}
\label{tab:selector-matched}
\small
\setlength{\tabcolsep}{4pt}
\renewcommand{\arraystretch}{1.12}
\begin{tabular}{@{}llrrrr@{}}
\toprule
Model & Dataset 
& Fixed $N=8$ Acc. 
& Adaptive Acc. 
& $\Delta$ Acc. 
& Sample red. \\
\midrule

\multirow{2}{*}{Phi-3-mini}
& MATH 
& 0.585 [0.558, 0.611] 
& 0.578 [0.552, 0.605] 
& $-0.007$ 
& 10.8\% \\

& GSM8K 
& 0.717 [0.692, 0.741] 
& 0.715 [0.690, 0.739] 
& $-0.002$ 
& 1.4\% \\

\midrule

\multirow{2}{*}{Qwen2.5-1.5B}
& MATH 
& 0.315 [0.290, 0.340] 
& 0.297 [0.273, 0.322] 
& $-0.018$ 
& 14.5\% \\

& GSM8K 
& 0.459 [0.432, 0.486] 
& 0.464 [0.437, 0.491] 
& $+0.005$ 
& 3.9\% \\

\bottomrule
\end{tabular}

\vspace{2pt}
\begin{minipage}{0.96\linewidth}
\footnotesize
Positive $\Delta$ Acc. means Adaptive is more accurate than fixed $N=8$. Negative values mean Adaptive has slightly lower accuracy while using fewer samples. The intervals are marginal Wilson intervals; paired tests require matched per-prompt correctness logs.
\end{minipage}
\end{table}

Adaptive is not always more accurate than fixed $N=8$. Instead, it gives an accuracy--compute tradeoff. On MATH, Adaptive uses fewer samples with only a small accuracy drop: 10.8\% fewer samples for Phi-3-mini and 14.5\% fewer samples for Qwen2.5-1.5B. On GSM8K, the accuracy is nearly unchanged, but the sample savings are smaller. With Qwen2.5-1.5B on GSM8K, Adaptive is slightly more accurate than fixed $N=8$ while using 3.9\% fewer samples.

These results support the main claim of the paper. The goal is not to show that adaptive budgeting always beats full-budget sampling in accuracy. The goal is to show that an interpretable controller can often stay close to full-budget accuracy while using fewer samples. This is most visible on MATH, where more prompts receive budgets below $N=8$.

\subsection{Interpreting The Accuracy--Compute Tradeoff}

The result should be read as a tradeoff, not only as an accuracy table. A method that always uses $N=8$ spends the maximum budget on every prompt. This can give strong accuracy, but it does not solve the allocation problem. The adaptive controller asks a different question: can we spend less on some prompts while keeping the final accuracy close?

\begin{figure}[h]
    \centering

    \begin{subfigure}[h]{0.48\textwidth}
        \centering
        \includegraphics[width=\linewidth]{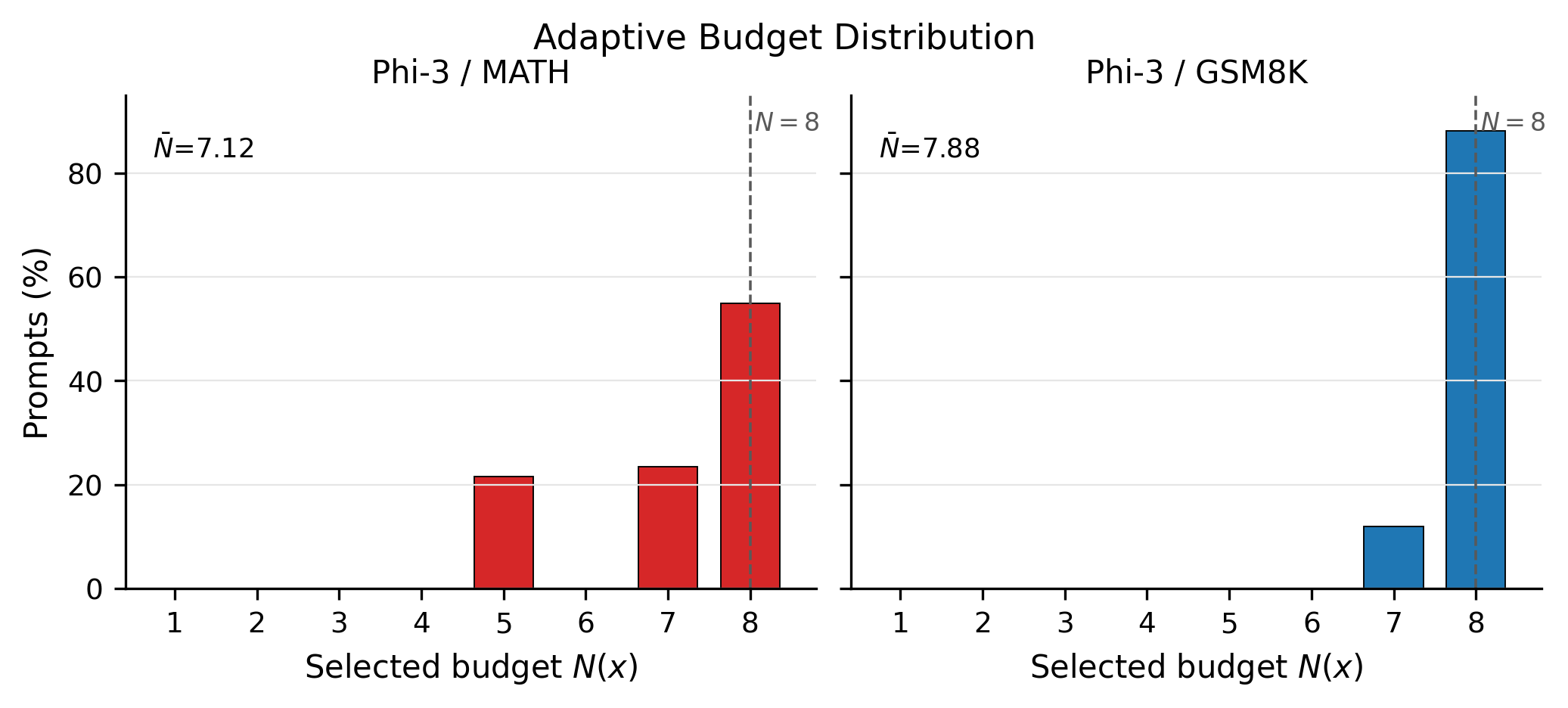}
        \caption{Phi-3-mini}
        \label{fig:budget-dist-phi3}
    \end{subfigure}
    \hfill
    \begin{subfigure}[h]{0.48\textwidth}
        \centering
        \includegraphics[width=\linewidth]{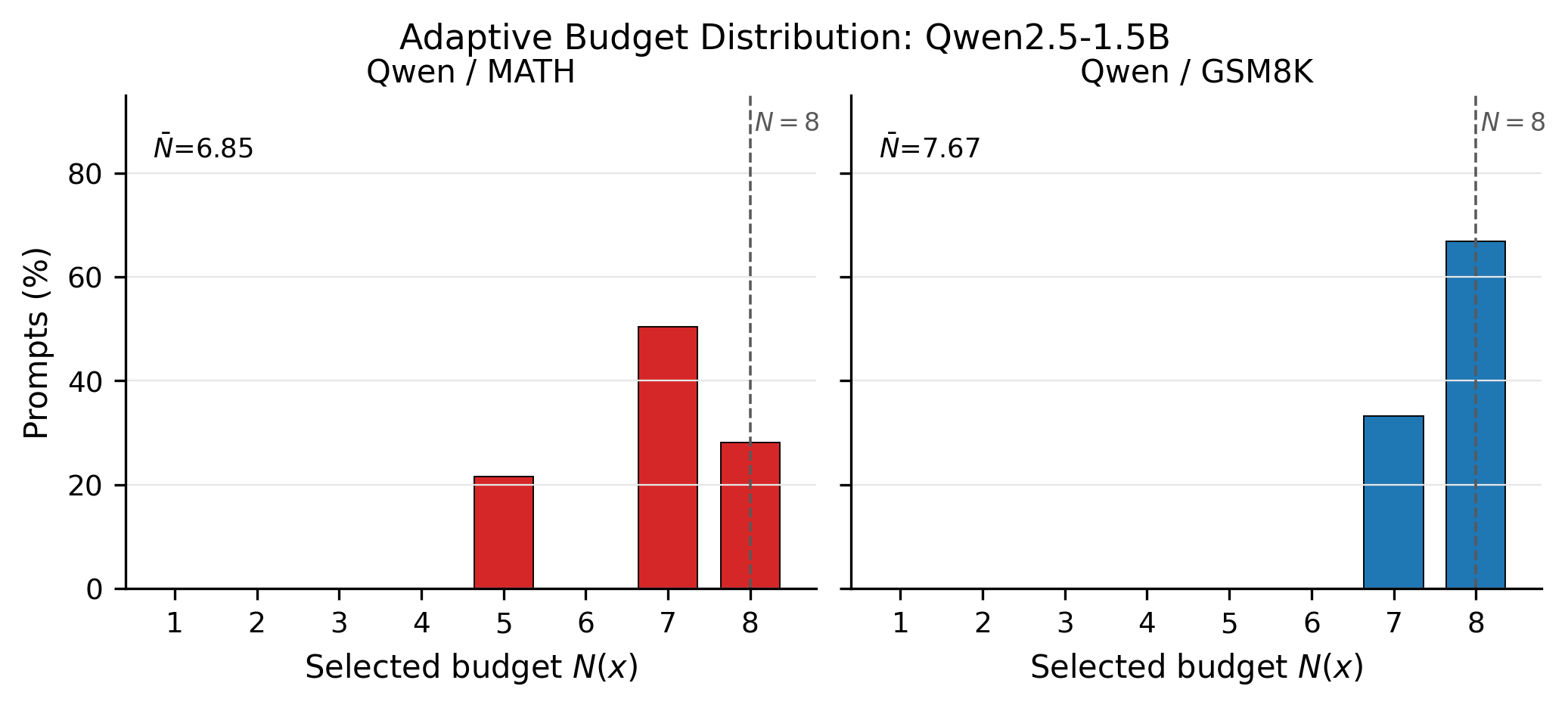}
        \caption{Qwen2.5-1.5B}
        \label{fig:budget-dist-qwen}
    \end{subfigure}

    \caption{Adaptive budget distributions under the fair-alignment setting. Each panel shows the percentage of prompts assigned to each sample budget \(N(x)\) for MATH and GSM8K. The dashed vertical line marks the full-budget baseline \(N=8\). For both models, MATH has more prompts assigned below \(N=8\), which explains the larger sample reductions on MATH. GSM8K is more concentrated near \(N=8\), showing that the controller behaves more conservatively on this dataset.}
    \label{fig:budget-distribution}
\end{figure}

Figure~\ref{fig:budget-distribution} helps explain the average sample counts in Table~\ref{tab:selector-matched}. The adaptive controller does not assign one fixed budget to every prompt. Instead, it produces different budget distributions depending on the model and dataset. On MATH, many prompts receive budgets below \(N=8\), so the average sample count decreases more. On GSM8K, most prompts receive \(N=8\), so the controller is more conservative. This suggests that many GSM8K prompts still benefit from near-full candidate aggregation. It also explains why sample savings are larger on MATH and smaller on GSM8K.

The broader comparison with standard baselines is reported in Table~\ref{tab:full-comparison} in the appendix. Adaptive improves over Best-of-$N$, compute-optimal allocation, and self-certainty-only baselines in the reported settings. The weaker performance of Best-of-3 and Best-of-5 shows that more samples alone are not enough. If the selector does not use the candidate set well, extra samples can add plausible wrong answers and reduce final accuracy. This is why the selector-matched fixed-$N=8$ baseline is important: it separates the effect of the budget policy from the effect of answer selection.

\subsection{Short Factual QA}

\begin{table}[h]
\centering
\caption{Fair-alignment results on SciQ with $n=1{,}000$ prompts.}
\label{tab:sciq}
\small
\setlength{\tabcolsep}{4pt}
\renewcommand{\arraystretch}{1.05}
\begin{tabular}{@{}lrrrr@{}}
\toprule
\multirow{2}{*}{Method} 
& \multicolumn{2}{c}{Phi-3-mini} 
& \multicolumn{2}{c}{Qwen2.5-1.5B} \\
\cmidrule(lr){2-3}\cmidrule(l){4-5}
& Acc. & Avg. $\bar N$ & Acc. & Avg. $\bar N$ \\
\midrule
Best-of-1\cite{snell2024scaling} & 0.153 & 1.00 & 0.365 & 1.00 \\
Best-of-3\cite{snell2024scaling} & 0.093 & 3.00 & 0.191 & 3.00 \\
Best-of-5 \cite{snell2024scaling}& 0.125 & 5.00 & 0.153 & 5.00 \\
Compute-optimal \cite{snell2024scaling} & 0.130 & 2.35 & 0.214 & 2.35 \\
Self-Certainty only \cite{borda} & 0.154 & 8.00 & 0.360 & 8.00 \\
Adaptive (ours) & \textbf{0.192} & 6.94 & \textbf{0.376} & 7.27 \\
\bottomrule
\end{tabular}
\end{table}

Table~\ref{tab:sciq} reports results on SciQ, a short factual question-answering task. Adaptive achieves the highest accuracy for both models. For Phi-3-mini, accuracy increases from 0.154 with self-certainty only to 0.192 with Adaptive, while the average budget decreases to 6.94 samples. For Qwen2.5-1.5B, Adaptive improves over Best-of-1 and self-certainty only, reaching 0.376 accuracy with 7.27 samples on average.

These results suggest that adaptive budgeting can also help outside math reasoning. The SciQ results should still be interpreted carefully because exact-match evaluation can be strict for short factual answers, especially for Phi-3-mini. Still, the pattern is useful: the adaptive controller improves accuracy while avoiding a fixed full-budget policy for every prompt.

\section{Ablation Study}
\label{sec:ablation}

Tables~\ref{tab:fixed-budget-ablation} and~\ref{tab:component-ablation} separate two ablation questions. Table~\ref{tab:fixed-budget-ablation} tests whether small fixed budgets are sufficient. Fixed caps of \(N=1,2,3\) perform much worse than Adaptive on both datasets, and fixed \(N=5\) is also lower. This shows that the system needs enough candidate answers for the selector to use agreement among samples.
Adaptive is close to fixed \(N=8\), but uses fewer samples. On GSM8K, Adaptive reaches 0.810 accuracy with an average of 7.86 samples, compared with 0.809 accuracy for fixed \(N=8\). On MATH, Adaptive reaches 0.570 accuracy with 7.07 samples, compared with 0.574 accuracy for fixed \(N=8\). This supports the main accuracy--compute tradeoff claim: the controller keeps most of the full-budget performance while reducing the average number of samples.

\begin{table}[h]
\centering
\caption{Fixed-budget controls on GSM8K and MATH with Phi-3-mini. Accuracy is exact-match accuracy.}
\label{tab:fixed-budget-ablation}
\small
\setlength{\tabcolsep}{4pt}
\renewcommand{\arraystretch}{1.08}
\begin{tabular}{@{}lrrrr@{}}
\toprule
\multirow{2}{*}{Variant} 
& \multicolumn{2}{c}{GSM8K} 
& \multicolumn{2}{c}{MATH} \\
\cmidrule(lr){2-3}\cmidrule(l){4-5}
& Acc. & Avg. $\bar N$ & Acc. & Avg. $\bar N$ \\
\midrule
Adaptive (ours) & 0.810 & 7.86 & 0.570 & 7.07 \\
Fixed adaptive $N=1$ & 0.725 & 1.00 & 0.457 & 1.00 \\
Fixed adaptive $N=2$ & 0.709 & 2.00 & 0.481 & 2.00 \\
Fixed adaptive $N=3$ & 0.712 & 3.00 & 0.456 & 3.00 \\
Fixed adaptive $N=5$ & 0.773 & 5.00 & 0.543 & 5.00 \\
Fixed adaptive $N=8$ & 0.809 & 8.00 & 0.574 & 8.00 \\
\bottomrule
\end{tabular}
\end{table}

We next test how removing individual controller components affects accuracy and sample count. 
Table~\ref{tab:component-ablation} shows that the results are mixed. Some removals have little effect, 
while some even improve accuracy. For example, removing entropy improves MATH accuracy, and 
neutralizing confidence also improves MATH accuracy while forcing the average budget to 8.00. These 
results should not be read as evidence that every signal is necessary or individually optimal. Instead, 
they show that the current fuzzy controller should be interpreted as a transparent allocation policy rather 
than an optimized learned policy. The usefulness of each refinement signal depends on the dataset, model, 
and selector. The most consistent pattern is that aggressive sample reduction hurts, while conservative 
adaptive allocation can preserve accuracy with fewer samples.
\begin{table}[h]
\centering
\caption{Component ablations on GSM8K and MATH with Phi-3-mini. Accuracy is exact-match accuracy. Individual refinement signals have mixed effects.}
\label{tab:component-ablation}
\small
\setlength{\tabcolsep}{4pt}
\renewcommand{\arraystretch}{1.08}
\begin{tabular}{@{}lrrrr@{}}
\toprule
\multirow{2}{*}{Variant} 
& \multicolumn{2}{c}{GSM8K} 
& \multicolumn{2}{c}{MATH} \\
\cmidrule(lr){2-3}\cmidrule(l){4-5}
& Acc. & Avg. $\bar N$ & Acc. & Avg. $\bar N$ \\
\midrule
Adaptive (ours) & 0.810 & 7.86 & 0.570 & 7.07 \\
No NLP features & 0.807 & 6.90 & 0.570 & 6.97 \\
No prompt-type signal & 0.818 & 7.89 & 0.566 & 7.12 \\
Neutral confidence & 0.801 & 8.00 & 0.601 & 8.00 \\
No entropy signal & 0.814 & 7.84 & 0.585 & 7.10 \\
No history cache & 0.793 & 7.58 & 0.567 & 6.82 \\
No rule adaptation & 0.798 & 7.58 & 0.561 & 6.82 \\
No type-2 uncertainty & 0.800 & 8.00 & 0.578 & 7.35 \\
\bottomrule
\end{tabular}
\end{table}

\begin{figure}[h]
    \centering
 \includegraphics[width=1\linewidth]{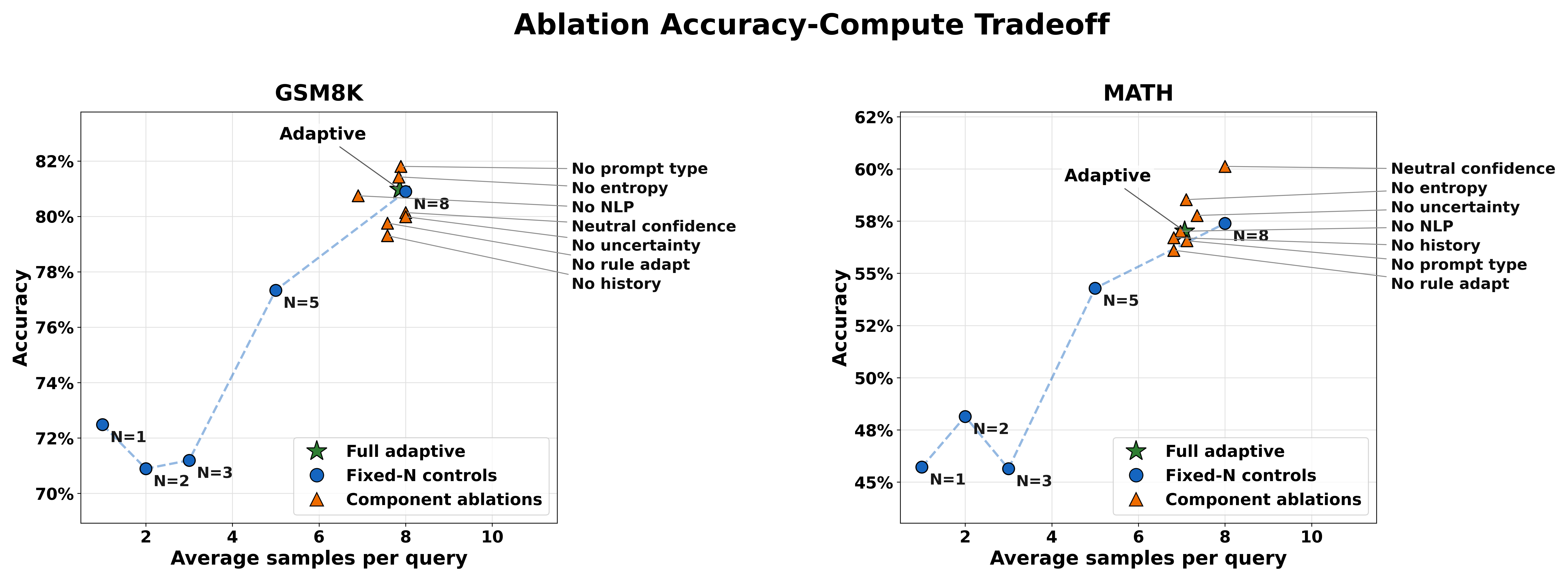}
    \caption{Ablation accuracy--compute tradeoff on GSM8K and MATH with Phi-3-mini. Each point is one ablation variant. Low fixed-budget controls reduce sample count but lose accuracy. Adaptive stays close to fixed \(N=8\) while using fewer samples, supporting the main accuracy--compute tradeoff claim.}
    \label{fig:ablation-tradeoff}
\end{figure}

Figure~\ref{fig:ablation-tradeoff} visualizes the full-set ablation results. Fixed low-budget variants move 
left because they use fewer samples, but they also lose clear accuracy. Adaptive remains near fixed 
\(N=8\), but with a smaller average budget. This supports a conservative adaptive strategy: the controller 
should reduce samples only when it can preserve the candidate diversity needed by the selector. Overall, 
the ablation results suggest that the main contribution is not that every hand-designed signal is individually 
optimal, but that an interpretable controller can provide a controllable accuracy--compute tradeoff. Future 
work can tune the fuzzy rules or learn signal weights while preserving interpretability.

\section{Limitations}
\label{sec:limitations}

The main limitation is that the fuzzy controller is hand-designed, so its rules and refinement weights are not guaranteed to be optimal for every model, dataset, or selector. The component ablations show that some signals can be helpful in one setting but neutral or harmful in another, suggesting that future versions should tune these weights more systematically while preserving interpretability. A second limitation is that answer selection remains a bottleneck: if a correct candidate is generated but not selected, better budgeting alone cannot fix the error. 
\section{Conclusion and Future Work}
\label{sec:conclusion}

We proposed an interpretable adaptive test-time scaling method that assigns a per-prompt sampling budget with a hierarchical fuzzy controller. The controller uses difficulty and uncertainty signals to decide how much inference-time computation each prompt should receive. Under fair-alignment settings, adaptive budgeting improves over standard fixed-budget and compute-allocation baselines and remains close to a selector-matched fixed-$N=8$ control while using fewer samples on average.  The results support the view that test-time scaling should not only increase compute, but allocate it more carefully. The controller gives a transparent way to decide when fewer samples are enough and when a prompt should stay near the full budget. This makes the allocation process easier to inspect than a fixed global budget or a black-box policy. Thus, fuzzy control provides a simple and practical direction for more efficient and transparent test-time reasoning. Future work should combine interpretable adaptive budgeting with stronger verifiers, learned rerankers, or dynamic early stopping, and test the method on broader tasks.

\appendix

\section{Appendix Overview}
The appendix contains details that are useful for reproducibility and interpretation. The main paper focuses on the question of whether adaptive budgeting improves the accuracy--compute tradeoff under fair comparison. The appendix gives the complete signal definitions, fuzzy-controller details, selector details, additional trace analyses, oracle audits, cost-saving analysis.

\subsection{Experiment Environment}
All experiments were run on a cluster using NVIDIA A100 GPUs (40GB), with standard multi-core CPU workers for data loading and preprocessing. Because running on the full dataset required substantially more compute time, we limited the number of runs/samples to keep the study feasible under available resources. Most individual runs used one A100 GPU and finished within a few hours, while larger robustness/ablation sweeps were parallelized across multiple GPUs and required longer wall-clock time.

\section{Complete Feature Definitions}
This section describes the nine controller signals in more detail. The surface complexity score $h$ combines length, punctuation, digits, uppercase ratio, and sentence count. Length is normalized by dividing the word count by $60$ and clipping to $[0,1]$. Punctuation and digits are included because mathematical and structured questions often contain symbols, numbers, or multi-clause instructions. Uppercase ratio is used only as a weak indicator because uppercase text can be noisy and should not dominate the score.

The NLP complexity score is designed to capture structure that surface features miss. Vocabulary richness measures whether the prompt uses varied language. Sentence count measures whether the prompt contains multiple clauses or steps. Math-symbol density captures formulas and arithmetic. Semantic ambiguity measures whether the prompt contains words or structures that can support multiple interpretations. Linguistic complexity captures syntactic load, while reasoning depth estimates how many reasoning steps the prompt may require. These components are combined into $c_{\mathrm{NLP}}$ using fixed weights. The weights are not learned; they are chosen to keep the score interpretable and stable.

Prompt type $\tau$ separates factual prompts from open-ended prompts. A short factual prompt usually needs less exploration, while an open-ended explanation or analysis prompt may benefit from more candidate diversity. Expected answer length $\lambda$ estimates the likely size of the final answer. This matters because longer answers have more opportunities for local errors and may need more samples to find a reliable candidate.

Confidence $\gamma$ and entropy $\eta$ are model-side signals. In the full system, they come from a short draft answer. Confidence is based on the geometric mean of token probabilities, while entropy measures how spread out the token distribution is. High confidence and low entropy usually indicate that the model has a stable answer. Low confidence and high entropy indicate uncertainty. The history signal $\pi$ is updated slowly after each prompt and stores a coarse estimate of how well the system performs on similar prompts. It is not meant to be a strong learned memory; it is a soft stabilizer.

\section{Detailed Fuzzy Controller}
The controller has two stages. The first stage performs the main budget decision. Complexity and confidence are each mapped to low, medium, and high membership functions. The rules are intuitive. Low complexity with high confidence activates low compute. Medium complexity or medium confidence activates medium compute. High complexity or low confidence activates high compute. Very high complexity and very low confidence activate very high compute.

The second stage applies refinements. Each refinement is small by design. High entropy increases compute. Long expected answer length increases compute. Open-ended prompt type increases exploration. Poor history increases compute because similar prompts have been difficult before. Semantic ambiguity, linguistic complexity, and reasoning depth each add a small positive adjustment. These refinements are represented as interval-valued adjustments, which makes the controller more conservative when signals disagree.

Defuzzification converts the activated intervals into a single scale. Let $a_j$ be the activation strength of output label $j$ and let $m_j$ be the center of its refined interval. The scale is computed as a weighted average,
\begin{equation}
 s = \frac{\sum_j a_jm_j}{\sum_j a_j+\epsilon},
\end{equation}
where $\epsilon$ prevents division by zero. The uncertainty value $u$ is computed from the average width of the active intervals. This uncertainty is used only to nudge the budget slightly upward; it does not override the scale.
\subsection{Supplementary Fuzzy Controller Specification}
\label{app:fuzzy_spec}

We use a deterministic Mamdani-style fuzzy controller with fixed membership functions, a fixed rule base, and centroid defuzzification. For each prompt \(x\), the controller maps normalized difficulty and uncertainty signals to a continuous sampling scale \(s(x)\in[0,1]\). This scale is later converted to an integer sample budget \(N(x)\).

\paragraph{Inputs and normalization.}
The primary controller inputs are prompt complexity \(c\), model confidence \(\gamma\), entropy \(\eta\), expected answer length \(\lambda\), prompt type \(\tau\), and history-cache score \(\pi\). All scalar inputs are normalized to \([0,1]\). For any raw scalar feature \(z_i\), we use min--max clipping:
\begin{equation}
\tilde{z}_i =
\mathrm{clip}\!\left(
\frac{z_i-a_i}{b_i-a_i},\,0,\,1
\right),
\end{equation}
where \((a_i,b_i)\) are estimated on a warm-up set and then kept fixed. Higher values of \(c\), \(\eta\), and \(\lambda\) indicate greater difficulty or uncertainty, while higher \(\gamma\) indicates greater model confidence. The history-cache score \(\pi\) represents prior success on similar prompts, so lower \(\pi\) increases the budget.

\paragraph{Membership functions.}
We use trapezoidal shoulder sets for boundary terms and triangular sets for middle terms. This avoids forcing the lowest and highest regions to peak at a single point.

The triangular membership function is
\begin{equation}
\mu_{\mathrm{tri}}(z;a,b,c)=
\begin{cases}
0, & z\le a \ \text{or}\ z\ge c,\\[2pt]
\dfrac{z-a}{b-a}, & a < z \le b,\\[6pt]
\dfrac{c-z}{c-b}, & b < z < c.
\end{cases}
\end{equation}

The trapezoidal membership function is
\begin{equation}
\mu_{\mathrm{trap}}(z;a,b,c,d)=
\begin{cases}
0, & z\le a \ \text{or}\ z\ge d,\\[2pt]
\dfrac{z-a}{b-a}, & a < z \le b,\\[6pt]
1, & b < z \le c,\\[2pt]
\dfrac{d-z}{d-c}, & c < z < d.
\end{cases}
\end{equation}

For each normalized input \(\tilde{z}\in[0,1]\), the fuzzy sets are
\begin{align}
L &= \mathrm{trap}(0.00,\,0.00,\,0.20,\,0.50),\\
M &= \mathrm{tri}(0.20,\,0.50,\,0.80),\\
H &= \mathrm{trap}(0.50,\,0.80,\,1.00,\,1.00).
\end{align}

Equivalently,
\begin{align}
\mu_L(\tilde{z}) &= \mu_{\mathrm{trap}}(\tilde{z};\,0.00,\,0.00,\,0.20,\,0.50),\\
\mu_M(\tilde{z}) &= \mu_{\mathrm{tri}}(\tilde{z};\,0.20,\,0.50,\,0.80),\\
\mu_H(\tilde{z}) &= \mu_{\mathrm{trap}}(\tilde{z};\,0.50,\,0.80,\,1.00,\,1.00).
\end{align}

The output variable is the sampling scale \(s(x)\). It uses five fuzzy sets:
\begin{align}
VL &= \mathrm{trap}(0.00,\,0.00,\,0.10,\,0.25),\\
L  &= \mathrm{tri}(0.10,\,0.25,\,0.45),\\
M  &= \mathrm{tri}(0.30,\,0.50,\,0.70),\\
H  &= \mathrm{tri}(0.55,\,0.75,\,0.90),\\
VH &= \mathrm{trap}(0.75,\,0.90,\,1.00,\,1.00).
\end{align}

\paragraph{Rule base.}
The first-stage rule base uses complexity \(c\) and confidence \(\gamma\). High complexity or low confidence increases the sampling scale, while low complexity and high confidence decrease it:
\begin{enumerate}
    \item If \(c\) is L and \(\gamma\) is H, then \(s\) is VL.
    \item If \(c\) is L and \(\gamma\) is M, then \(s\) is L.
    \item If \(c\) is L and \(\gamma\) is L, then \(s\) is M.
    \item If \(c\) is M and \(\gamma\) is H, then \(s\) is L.
    \item If \(c\) is M and \(\gamma\) is M, then \(s\) is M.
    \item If \(c\) is M and \(\gamma\) is L, then \(s\) is H.
    \item If \(c\) is H and \(\gamma\) is H, then \(s\) is M.
    \item If \(c\) is H and \(\gamma\) is M, then \(s\) is H.
    \item If \(c\) is H and \(\gamma\) is L, then \(s\) is VH.
\end{enumerate}

\paragraph{Second-stage refinements.}
The first-stage output is refined using entropy \(\eta\), expected answer length \(\lambda\), prompt type \(\tau\), history-cache score \(\pi\), semantic ambiguity \(a_{\mathrm{sem}}\), linguistic complexity \(a_{\mathrm{ling}}\), and reasoning-depth score \(a_{\mathrm{rea}}\). Each refinement is small by design:
\begin{equation}
\Delta =
0.10\eta
+0.08\lambda
+0.05\tau
+0.08(1-\pi)
+0.05a_{\mathrm{sem}}
+0.04a_{\mathrm{ling}}
+0.05a_{\mathrm{rea}} .
\end{equation}
The refined scale is
\begin{equation}
s'(x)=\mathrm{clip}(s(x)+\Delta,\,0,\,1).
\end{equation}

\paragraph{Inference and defuzzification.}
We use product \(t\)-norm for AND, max aggregation across activated rules, and centroid defuzzification on a uniform 1000-point grid over \([0,1]\). The rule base is fixed and is not learned or tuned during evaluation.

\paragraph{Controller uncertainty.}
To estimate controller uncertainty, we compute the average width of the active output regions after refinement. This gives an uncertainty value \(u(x)\in[0,1]\). The scale is then nudged upward when the controller is uncertain:
\begin{equation}
\bar{s}(x)=\mathrm{clip}\bigl(s'(x)+\alpha u(x),\,0,\,1\bigr),
\quad \alpha=0.2.
\end{equation}

\paragraph{Budget mapping.}
Finally, the continuous scale is converted into an integer sample budget:
\begin{equation}
N(x)=
\max\left(1,\mathrm{round}\left(1+\bar{s}(x)(N_{\max}-1)\right)\right).
\end{equation}
For clearly difficult prompts, we apply a conservative floor:
\begin{equation}
\text{if } c>0.6 \text{ and } N_{\max}\ge 6,\quad
N(x)\leftarrow \max\bigl(N(x),\min(6,N_{\max})\bigr).
\end{equation}
This prevents prompts with high estimated difficulty from receiving very small budgets.

\section{Selector Details}
The selector is intentionally kept separate from the budget controller. For each candidate answer, the system computes a self-certainty score from token-level log probabilities. Candidates are sorted by this score. The final answer string is extracted from each candidate using the same extraction routine across all methods. Borda aggregation then assigns more points to answers supported by higher-ranked candidates. If several candidates produce the same final answer, their Borda scores accumulate. The final answer is the extracted answer string with the largest total score.

This selector is useful because it combines two kinds of information. Self-certainty measures how internally stable each candidate is, while Borda aggregation measures agreement among candidates. A single candidate with high self-certainty can still lose if many other candidates agree on a different final answer. This is important in math tasks, where different reasoning paths can lead to the same final value.

For fair evaluation, any fixed-budget baseline used to isolate budgeting should use the same selector. Otherwise, the comparison mixes two effects: budget allocation and answer selection. 

\section{Additional Trace Diagnostics}
\label{app:additional-trace-diagnostics}

This appendix provides additional trace-level diagnostics for the adaptive GSM8K run with Phi-3-medium and \(N_{\max}=16\). These figures are not part of the main fair-alignment comparison. They are included to show how the controller behaves inside a full adaptive run. The diagnostics show how the selected sample budget changes across prompts, how the history-cache signal evolves, how correctness is distributed across selected budgets, and how controller signals relate to budget decisions.
\begin{figure*}[h]
    \centering

    \begin{subfigure}[h]{0.48\textwidth}
        \centering
        \includegraphics[width=\linewidth]{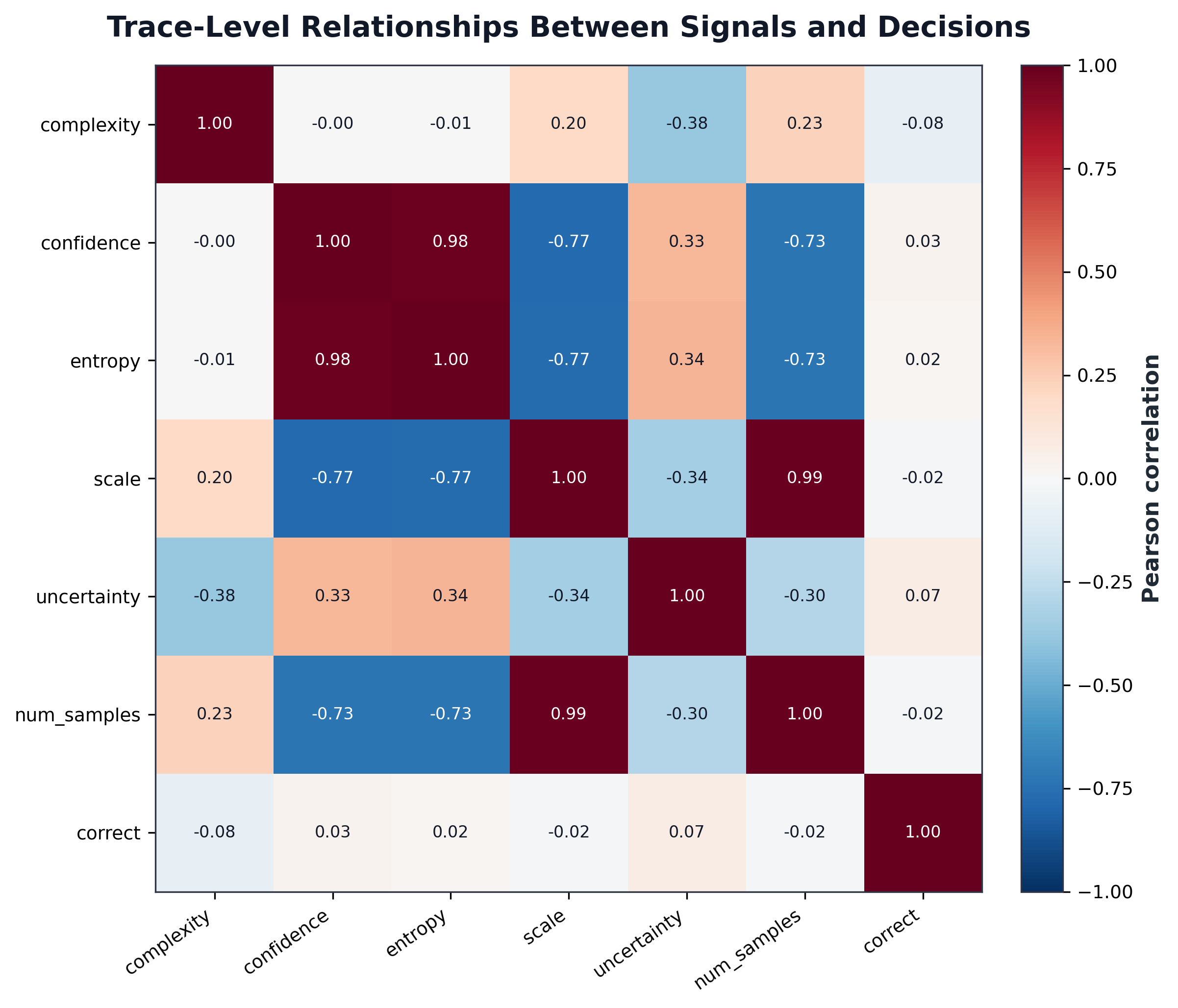}
        \caption{Signal correlations.}
        \label{fig:trace-correlation}
    \end{subfigure}
    \hfill
    \begin{subfigure}[h]{0.48\textwidth}
        \centering
        \includegraphics[width=\linewidth]{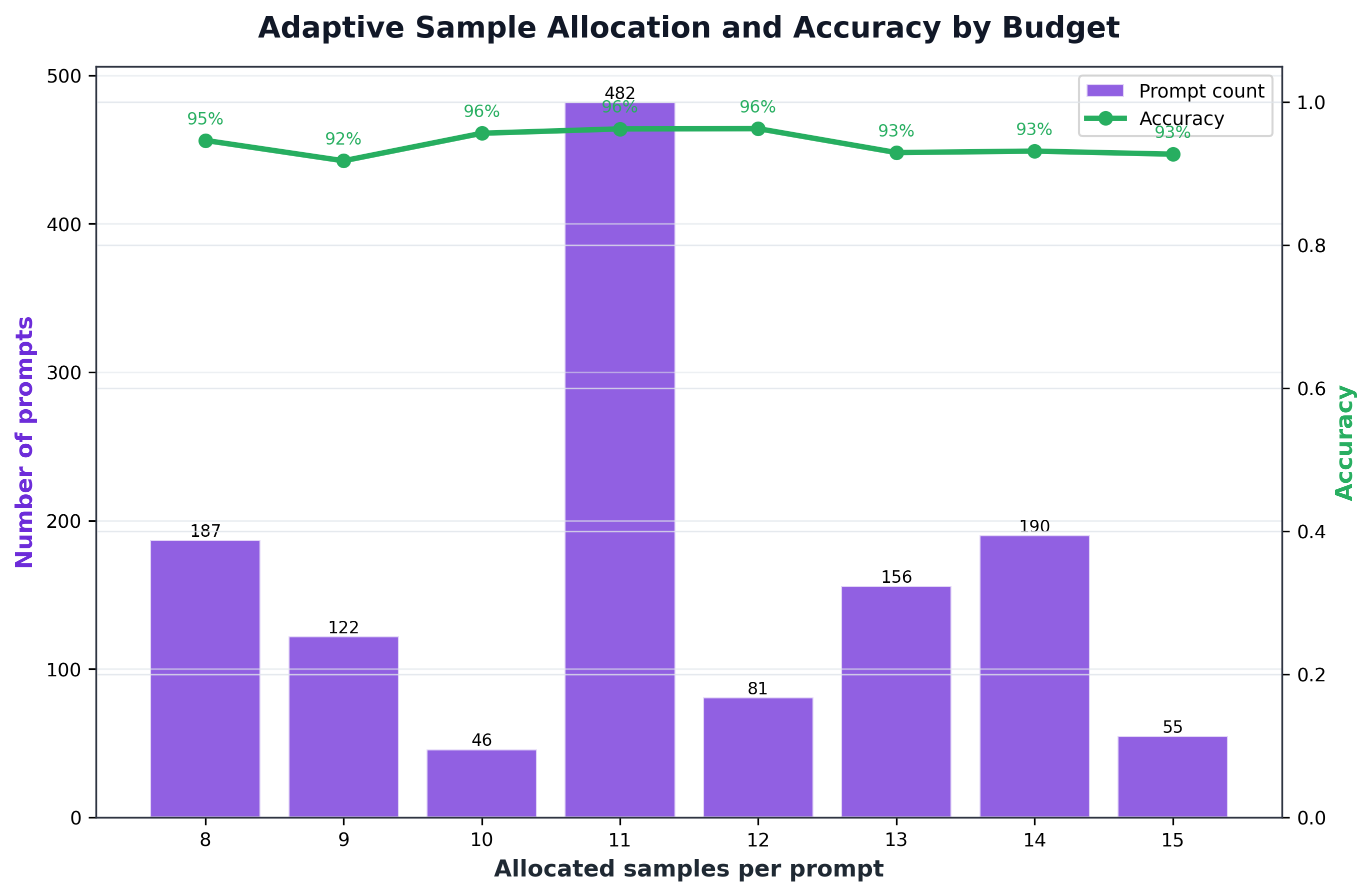}
        \caption{Budget allocation and accuracy.}
        \label{fig:trace-budget-accuracy}
    \end{subfigure}

    \vspace{2mm}

    \begin{subfigure}[h]{0.48\textwidth}
        \centering
        \includegraphics[width=\linewidth]{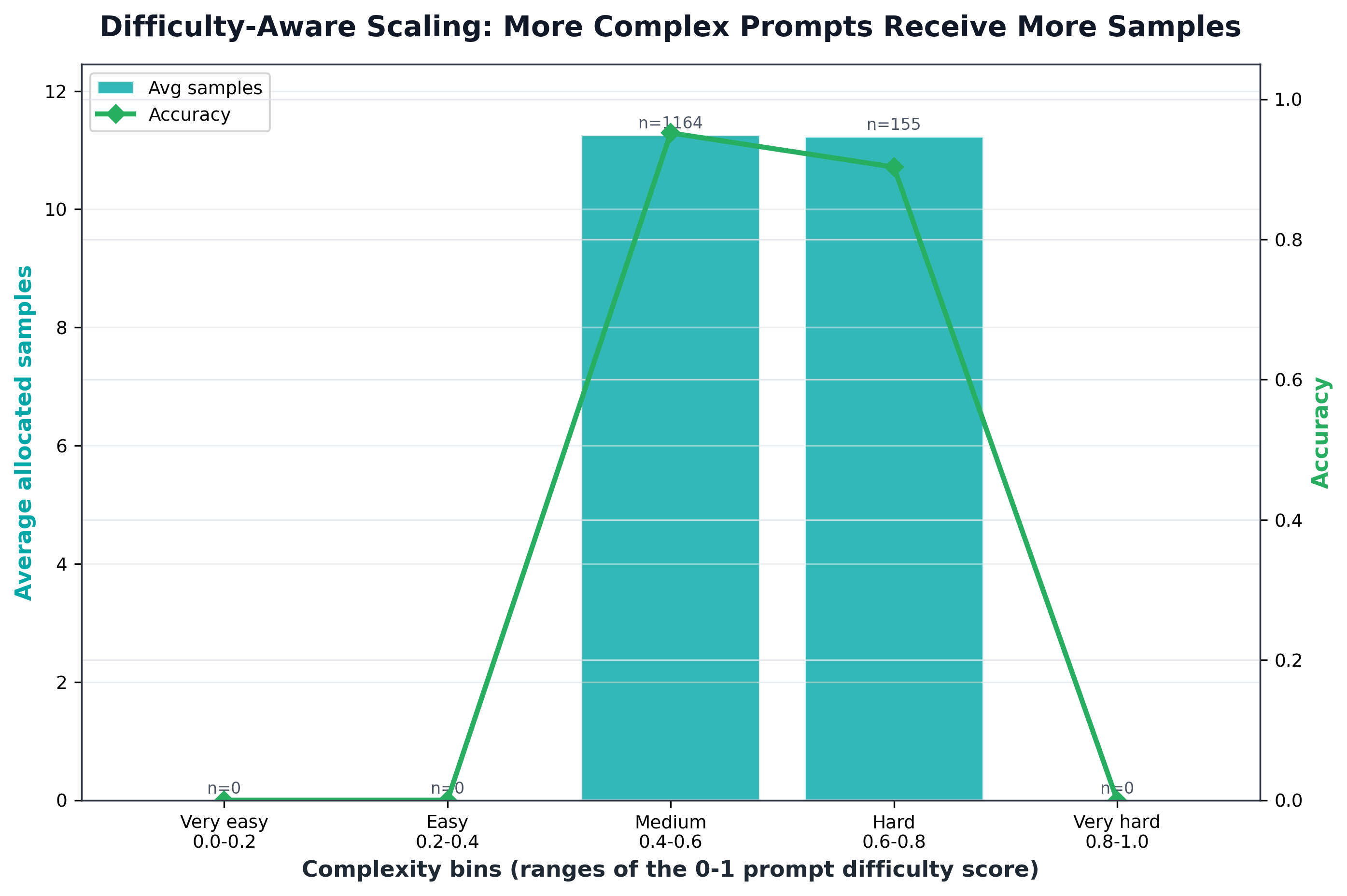}
        \caption{Complexity bins.}
        \label{fig:trace-complexity}
    \end{subfigure}
    \hfill
    \begin{subfigure}[h]{0.48\textwidth}
        \centering
        \includegraphics[width=\linewidth]{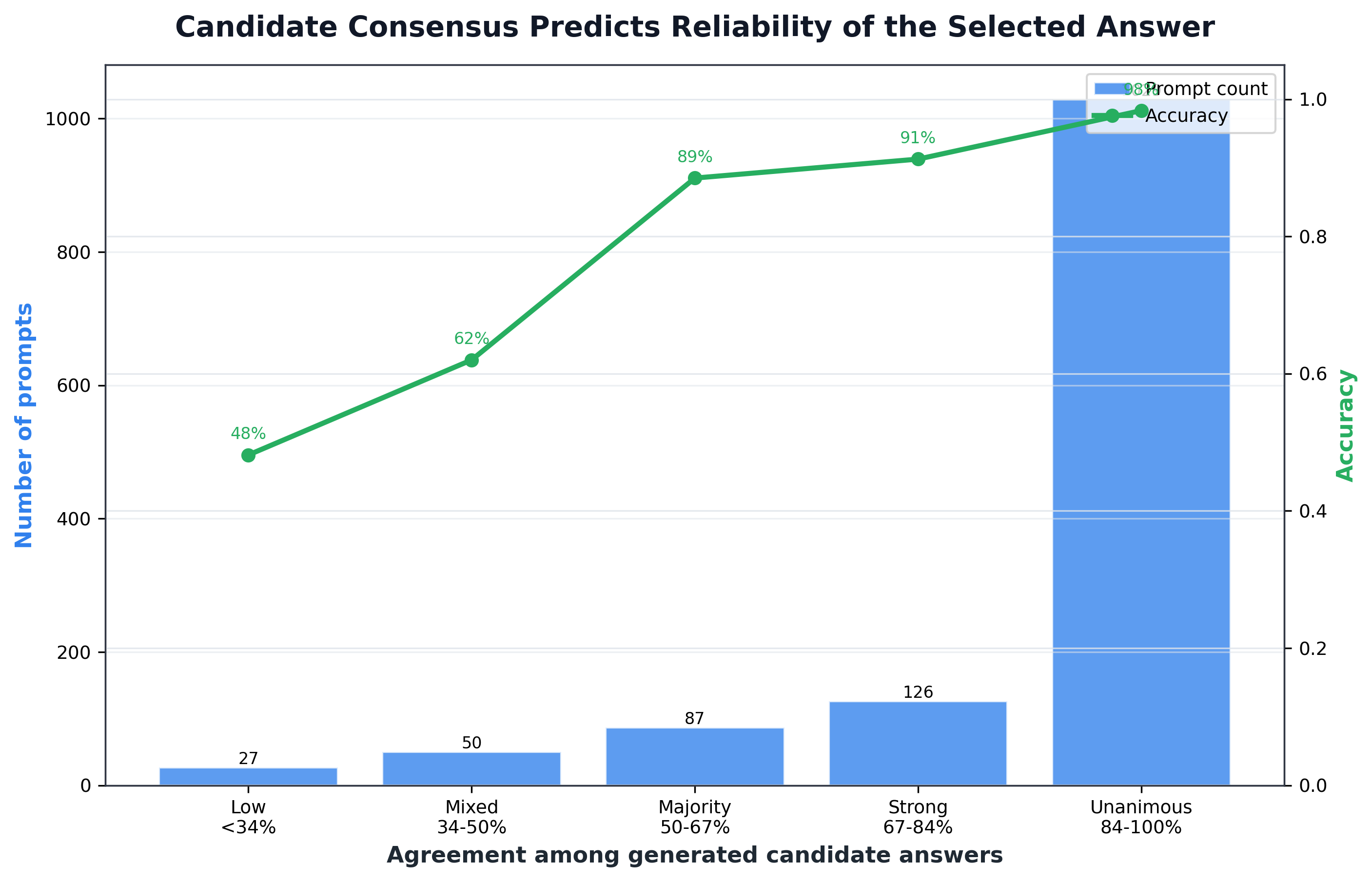}
        \caption{Candidate consensus.}
        \label{fig:trace-consensus}
    \end{subfigure}

    \caption{Trace-level diagnostics for the adaptive GSM8K run with Phi-3-medium and \(N_{\max}=16\). The plots show how controller signals relate to budget decisions and final accuracy. The sample count is closely tied to the controller scale, higher-complexity prompts receive larger budgets, and stronger candidate agreement is associated with higher accuracy.}
    \label{fig:trace-diagnostics}
\end{figure*}

We also conducted tests for larger values of \( N \) and report trace-level diagnostics for the adaptive GSM8K run with Phi-3-medium and \( N_{\max} = 32 \). These results are diagnostic and are not part of the main fair-alignment comparison. They provide insight into the controller's behavior when a larger maximum sample budget is available.

\begin{figure}[h]
    \centering

    \begin{subfigure}[h]{0.48\textwidth}
        \centering
        \includegraphics[width=\linewidth]{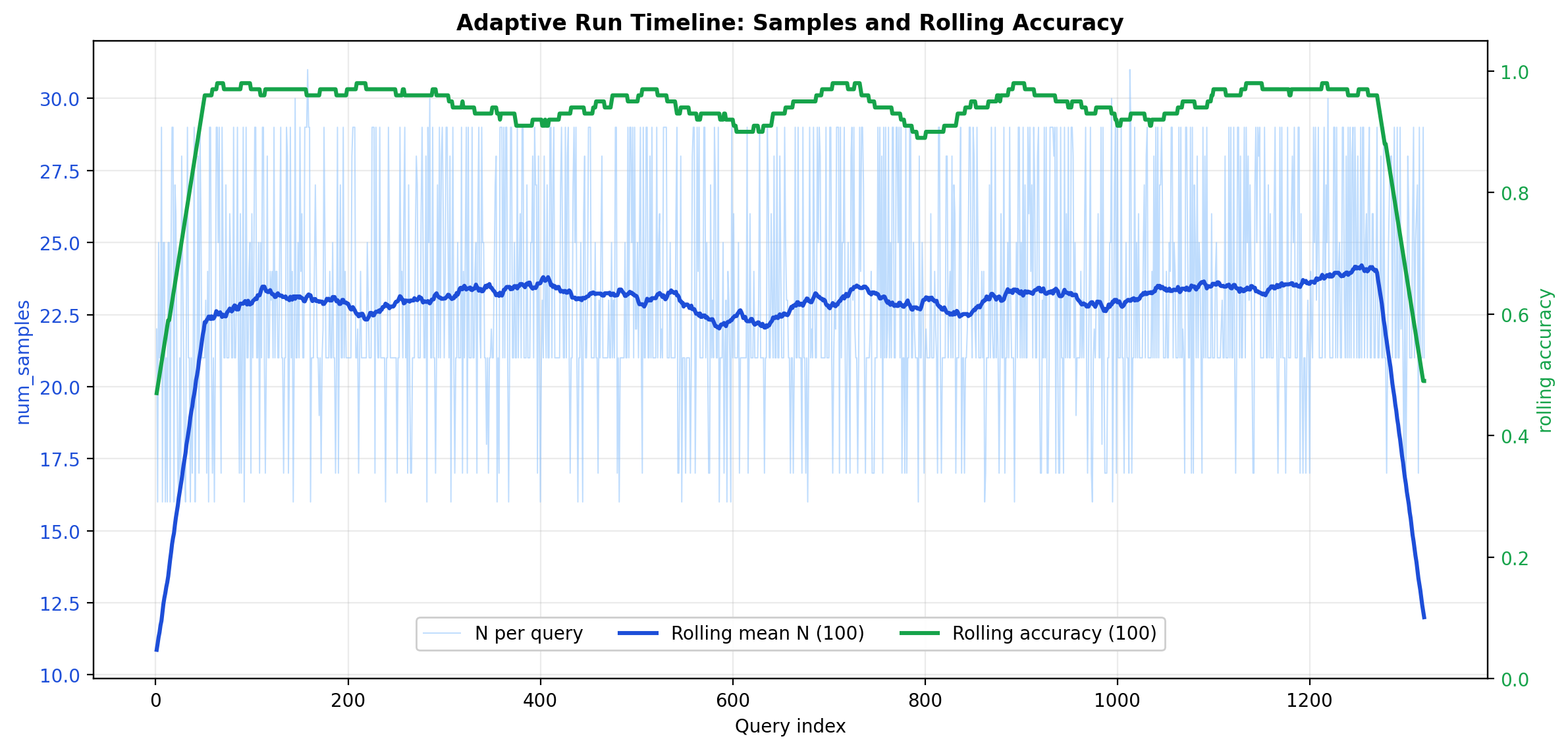}
        \caption{Samples and rolling accuracy over the run.}
        \label{fig:trace-timeline-n32}
    \end{subfigure}
    \hfill
    \begin{subfigure}[h]{0.48\textwidth}
        \centering
        \includegraphics[width=\linewidth]{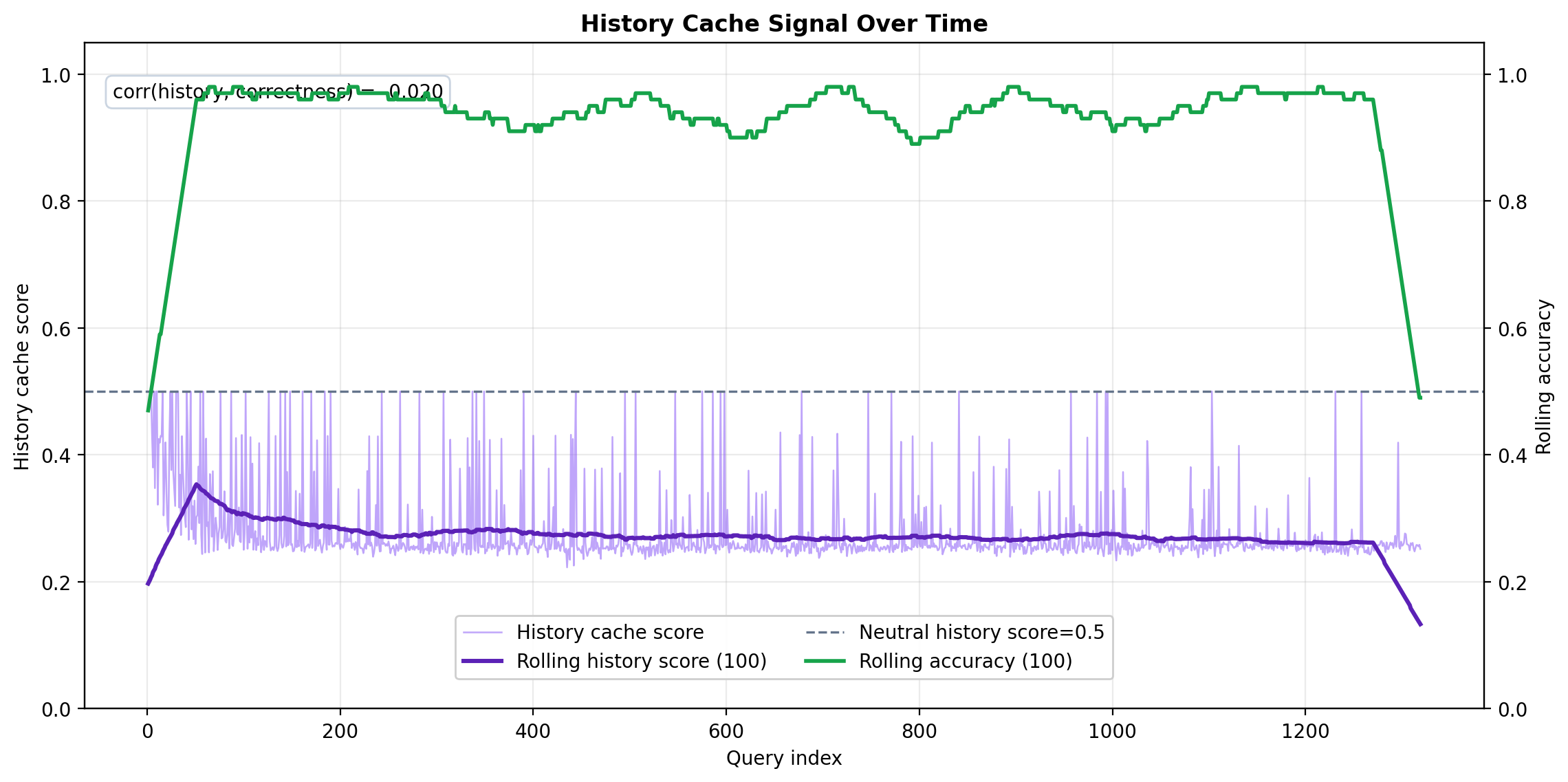}
        \caption{History-cache signal over time.}
        \label{fig:trace-history-n32}
    \end{subfigure}

    \vspace{2mm}

    \begin{subfigure}[h]{0.62\textwidth}
        \centering
        \includegraphics[width=\linewidth]{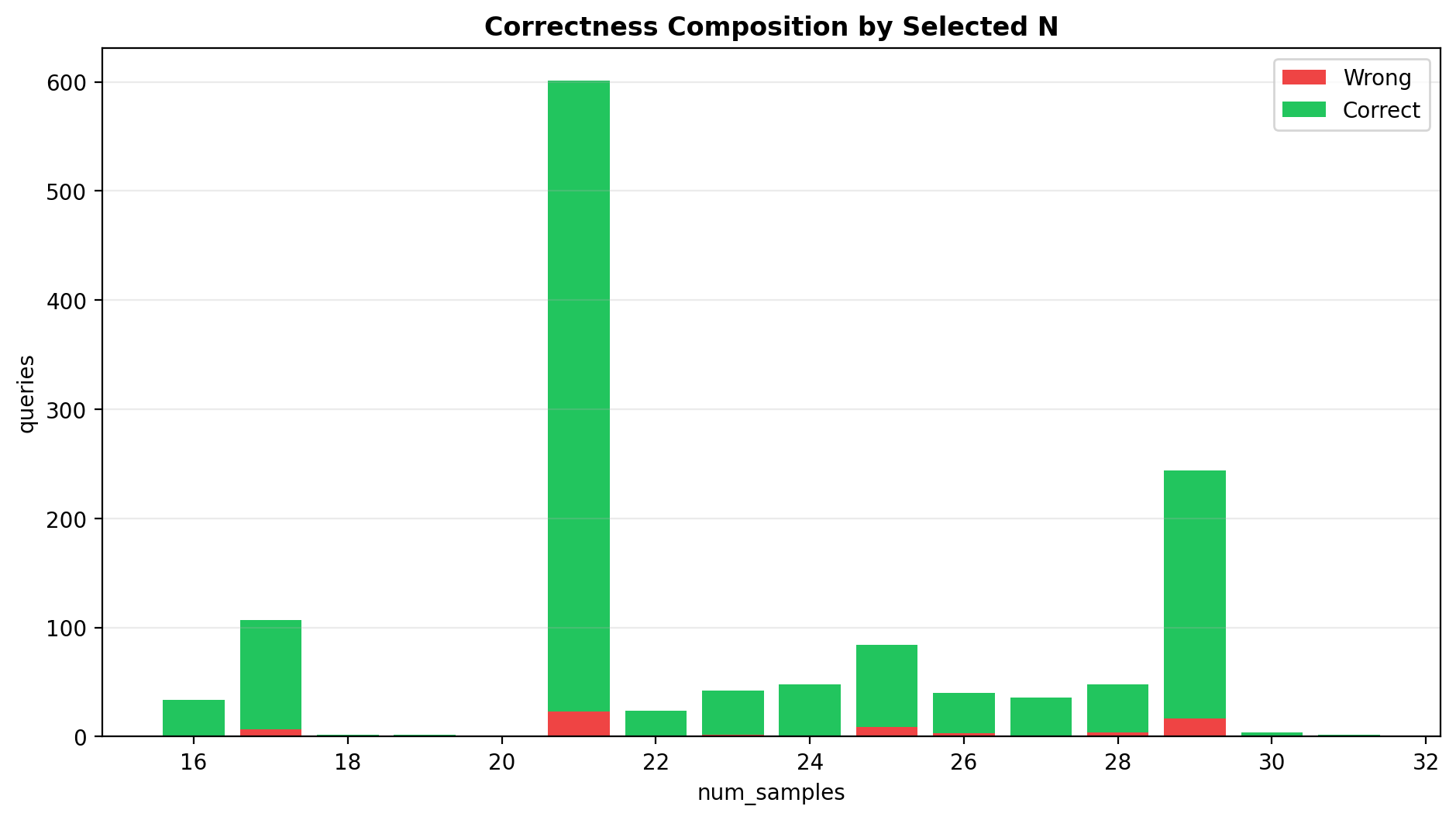}
        \caption{Correctness composition by selected sample count.}
        \label{fig:trace-correctness-stack-n32}
    \end{subfigure}

    \caption{Additional run-level diagnostics for the adaptive GSM8K run with Phi-3-medium and \(N_{\max}=32\).}
    \label{fig:additional-trace-diagnostics-n32}
\end{figure}

Figure~\ref{fig:additional-trace-diagnostics-n32} gives run-level diagnostics. The timeline shows that sample counts vary across queries while rolling accuracy stays high for most of the run. The history-cache signal changes over time, but its direct relationship with correctness is weak. The correctness composition plot shows that most budget groups contain many more correct than incorrect predictions.

\subsection{Supplementary GSM8K Adaptive-Only Runs at Larger Budgets}
\label{app:gsm8k-adaptive-n16-n32}

This appendix reports an additional diagnostic experiment on GSM8K. These runs are \emph{not} part of the main fair-alignment protocol. In the main experiments, decoding settings are fixed and the adaptive controller changes only the number of samples. Here, we instead run the full adaptive pipeline: the controller uses a draft pass to estimate model-side signals, and the decoding parameters are allowed to vary with the selected budget. Therefore, the results in this section should be interpreted as a high-budget behavior analysis, not as a direct comparison to the fixed-budget baselines in the main tables.

We evaluate on the full GSM8K test split with $n=1{,}319$ examples. The runs use two open instruction-tuned models: Microsoft Phi-3-medium-4k-instruct and Qwen2.5-1.5B-Instruct. For each model, we test two maximum sample caps, $N_{\max}\in\{16,32\}$. All runs use self-certainty ranking with Borda aggregation over extracted final answers.

Table~\ref{tab:app-gsm8k-adaptive-n16-n32} reports exact-match accuracy, average selected samples $\bar N$, oracle accuracy, selection-miss rate, and average logged latency. Oracle accuracy is the fraction of prompts for which at least one generated candidate matches the gold answer. The selection-miss rate is the fraction of prompts where a correct candidate exists but the selector returns an incorrect final answer. This audit separates the quality of the generated candidate pool from the quality of the final selector.

\begin{table}[h]
  \centering
  \small
  \caption{Adaptive-only GSM8K results at larger sample caps. These runs use the full adaptive pipeline and adaptive decoding, so they are diagnostic and not directly comparable to the fair-alignment results in the main text.}
  \label{tab:app-gsm8k-adaptive-n16-n32}
  \begin{tabular}{@{}lrrrrr@{}}
    \toprule
    Model & $N_{\max}$ & Accuracy & $\bar N$ & Oracle acc. & Sel.\ miss \\
    \midrule
    Phi-3-medium & 16 & 0.946 & 11.25 & 0.980 & 3.34\% \\
    Phi-3-medium & 32 & 0.948 & 23.08 & 0.986 & 3.71\% \\
    Qwen2.5-1.5B & 16 & 0.571 & 13.70 & 0.820 & 24.94\% \\
    Qwen2.5-1.5B & 32 & 0.645 & 27.28 & 0.887 & 24.18\% \\
    \bottomrule
  \end{tabular}
\end{table}

The results show two different regimes. For Phi-3-medium, increasing $N_{\max}$ from 16 to 32 gives only a small accuracy gain, from 0.946 to 0.948, while roughly doubling the average number of samples. The oracle accuracy is already very high at $N_{\max}=16$, and the selection-miss rate remains low. This suggests that, for this stronger model, the selector is already close to the best answer available in the candidate pool, and additional samples provide limited marginal benefit.

The Qwen2.5-1.5B results show a different pattern. Increasing $N_{\max}$ from 16 to 32 improves accuracy from 0.571 to 0.645. However, the oracle accuracy is much higher than the selected accuracy in both cases. At $N_{\max}=32$, the candidate pool contains a correct answer for 0.887 of prompts, but the selected answer is correct for only 0.645 of prompts. The selection-miss rate remains around 24--25\%. This indicates that the smaller model is strongly selection-limited: the system often generates a correct candidate, but the current selector does not always choose it.

These diagnostic runs support the main paper's interpretation that budget allocation and answer selection are separate bottlenecks. Larger budgets can improve the candidate pool, especially for smaller models, but better selection or verification is needed to fully use that pool. They also show why the main fair-alignment protocol is necessary: without fixed decoding and matched selectors, high-budget adaptive-only runs are useful for understanding behavior but should not be used as the primary baseline comparison.

\section{Cumulative Compute Savings}
\label{app:cumulative-compute}

This appendix compares the cumulative sample cost of adaptive sampling against a fixed maximum-budget policy. For each run, the red line shows the cumulative samples that would be used by always sampling at the maximum budget, and the blue line shows the cumulative samples used by the adaptive controller. The shaded area is the saved computation. These plots show that the controller reduces total sample usage across the full run rather than only on a small subset of prompts.

\begin{figure}[h]
    \centering

    \begin{subfigure}[h]{0.48\textwidth}
        \centering
        \includegraphics[width=\linewidth]{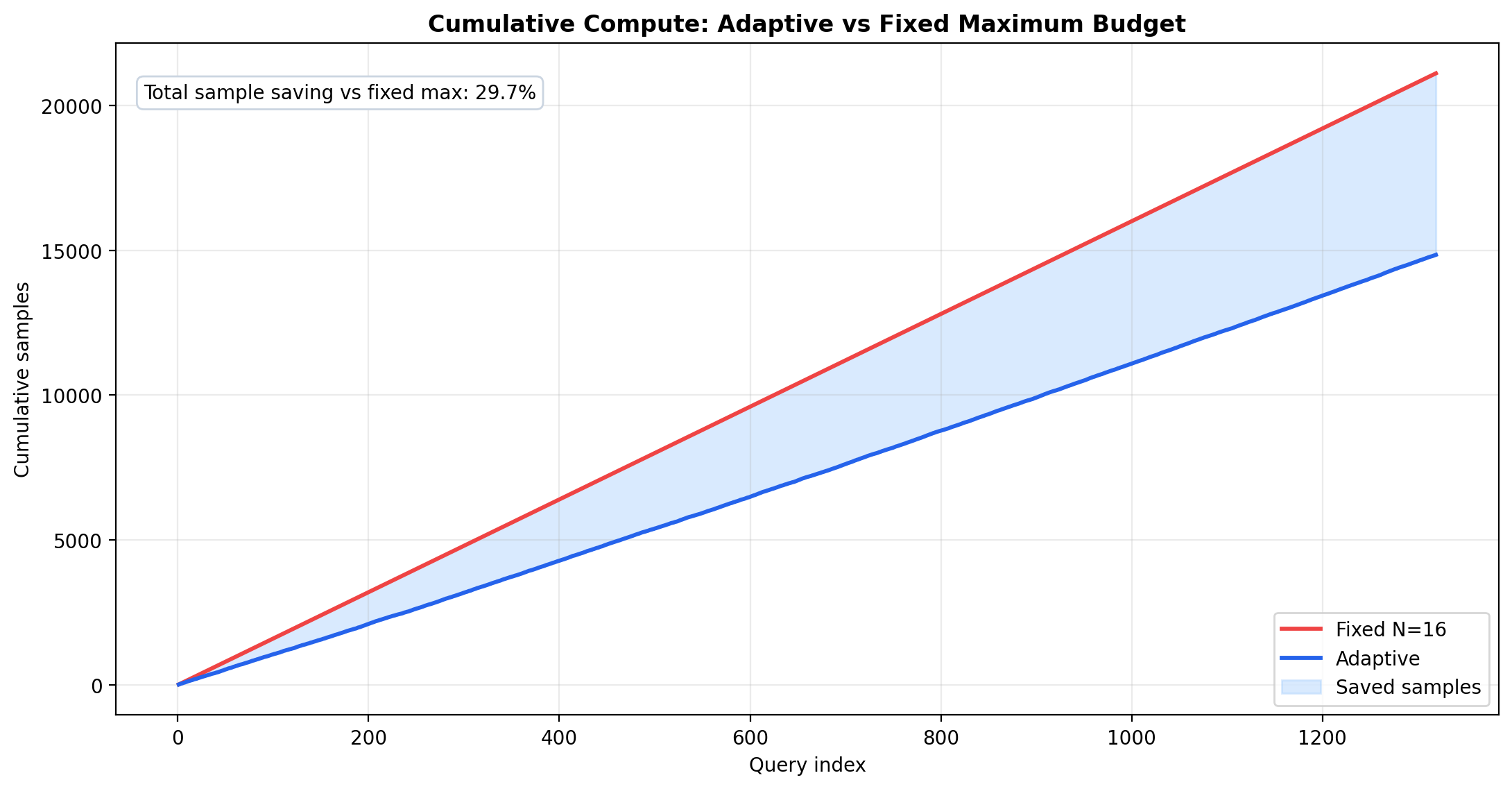}
        \caption{Phi-3-medium, \(N_{\max}=16\).}
        \label{fig:phi3-n16-compute}
    \end{subfigure}
    \hfill
    \begin{subfigure}[h]{0.48\textwidth}
        \centering
        \includegraphics[width=\linewidth]{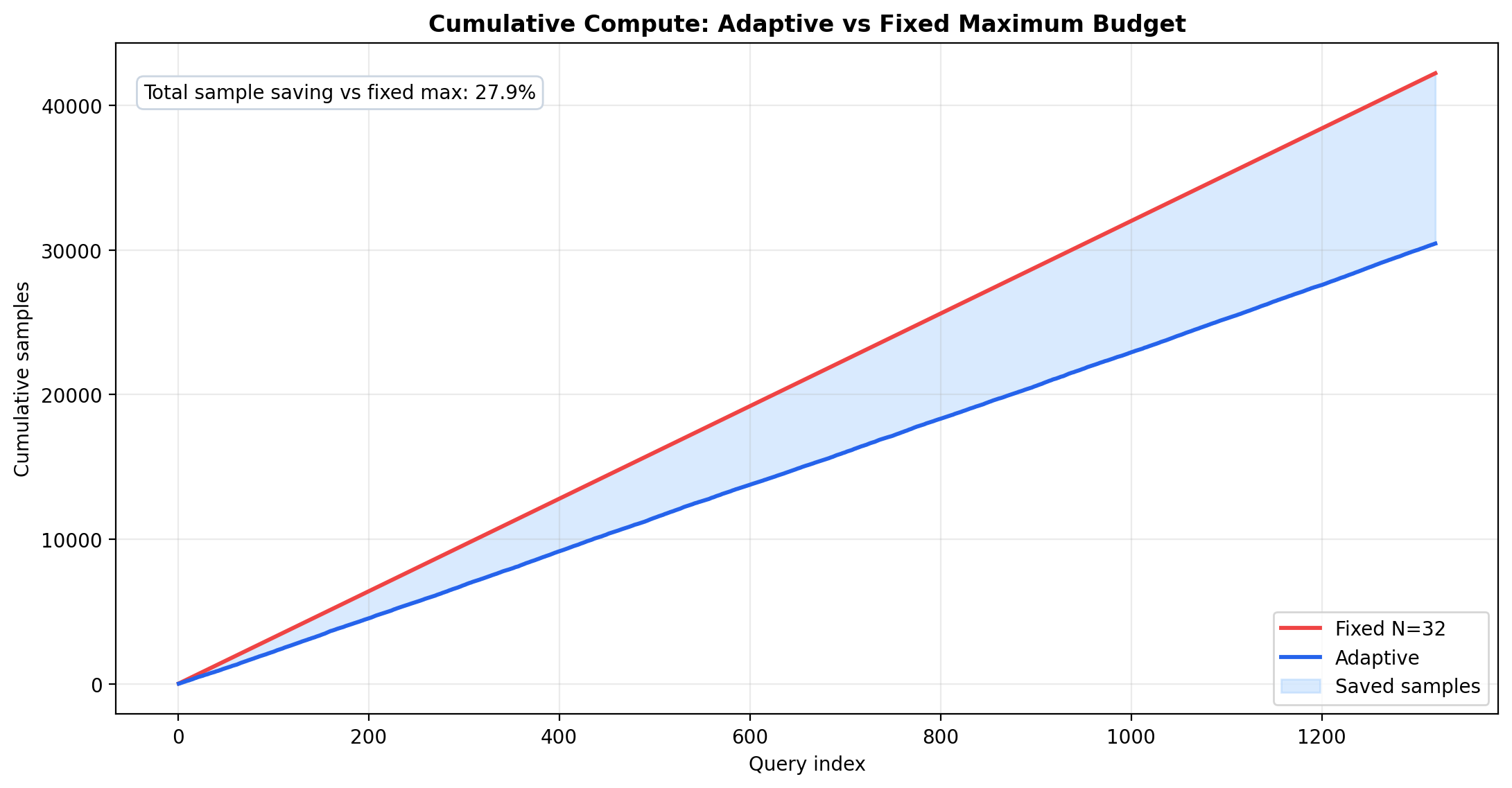}
        \caption{Phi-3-medium, \(N_{\max}=32\).}
        \label{fig:phi3-n32-compute}
    \end{subfigure}

    \caption{Cumulative compute savings for Phi-3-medium adaptive GSM8K runs. The adaptive controller saves \(29.7\%\) of samples relative to fixed \(N=16\) and \(27.9\%\) relative to fixed \(N=32\).}
    \label{fig:phi3-cumulative-compute}
\end{figure}

Figure~\ref{fig:phi3-cumulative-compute} shows that Phi-3-medium obtains consistent compute savings at both budget caps. The saved area grows steadily over the run, meaning that adaptive sampling reduces total computation throughout the evaluation rather than only at the beginning or end. The savings remain large even when the maximum budget increases from 16 to 32.

\begin{figure}[h]
    \centering

    \begin{subfigure}[h]{0.48\textwidth}
        \centering
        \includegraphics[width=\linewidth]{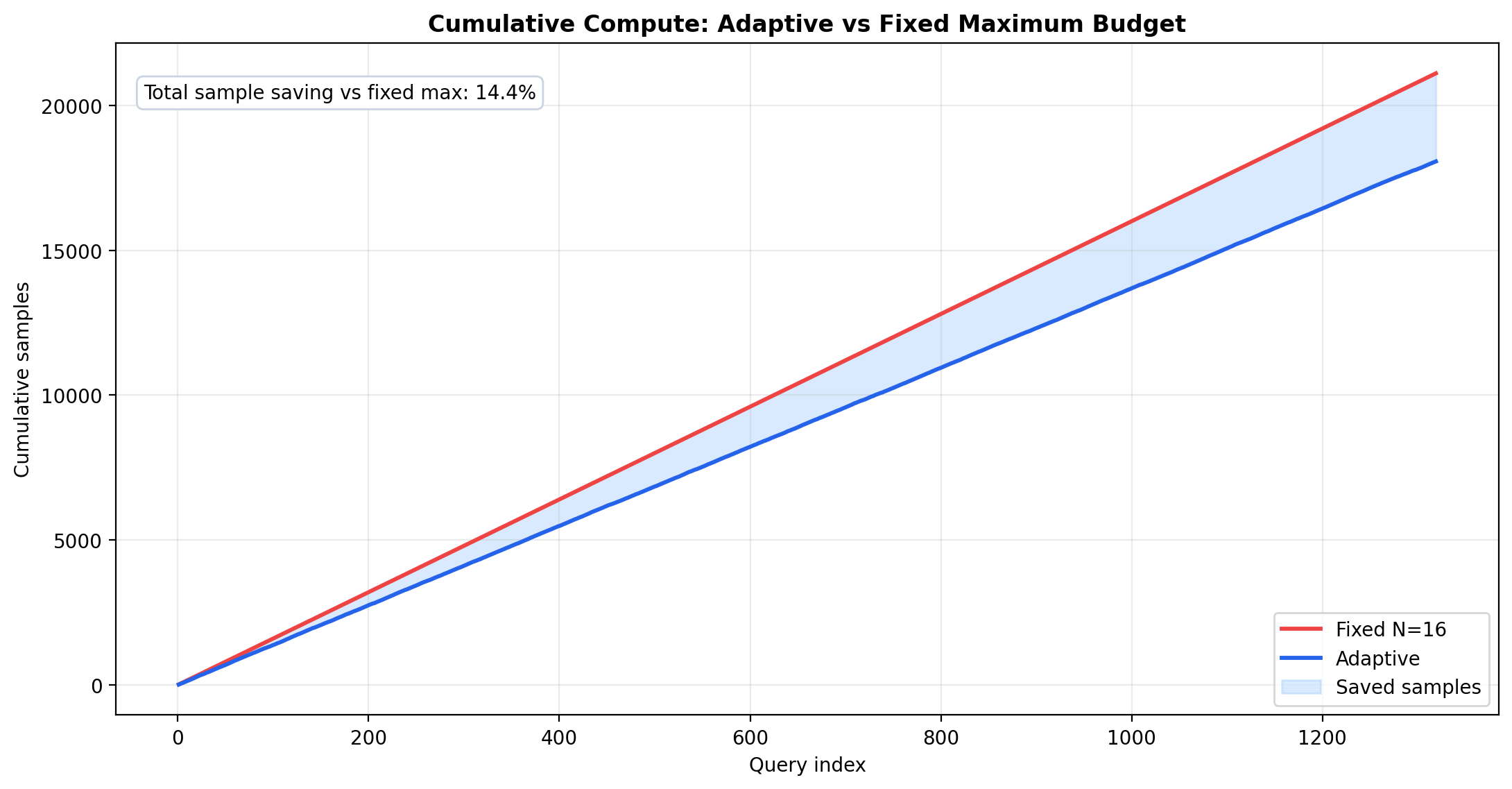}
        \caption{Qwen2.5-1.5B, \(N_{\max}=16\).}
        \label{fig:qwen-n16-compute}
    \end{subfigure}
    \hfill
    \begin{subfigure}[h]{0.48\textwidth}
        \centering
        \includegraphics[width=\linewidth]{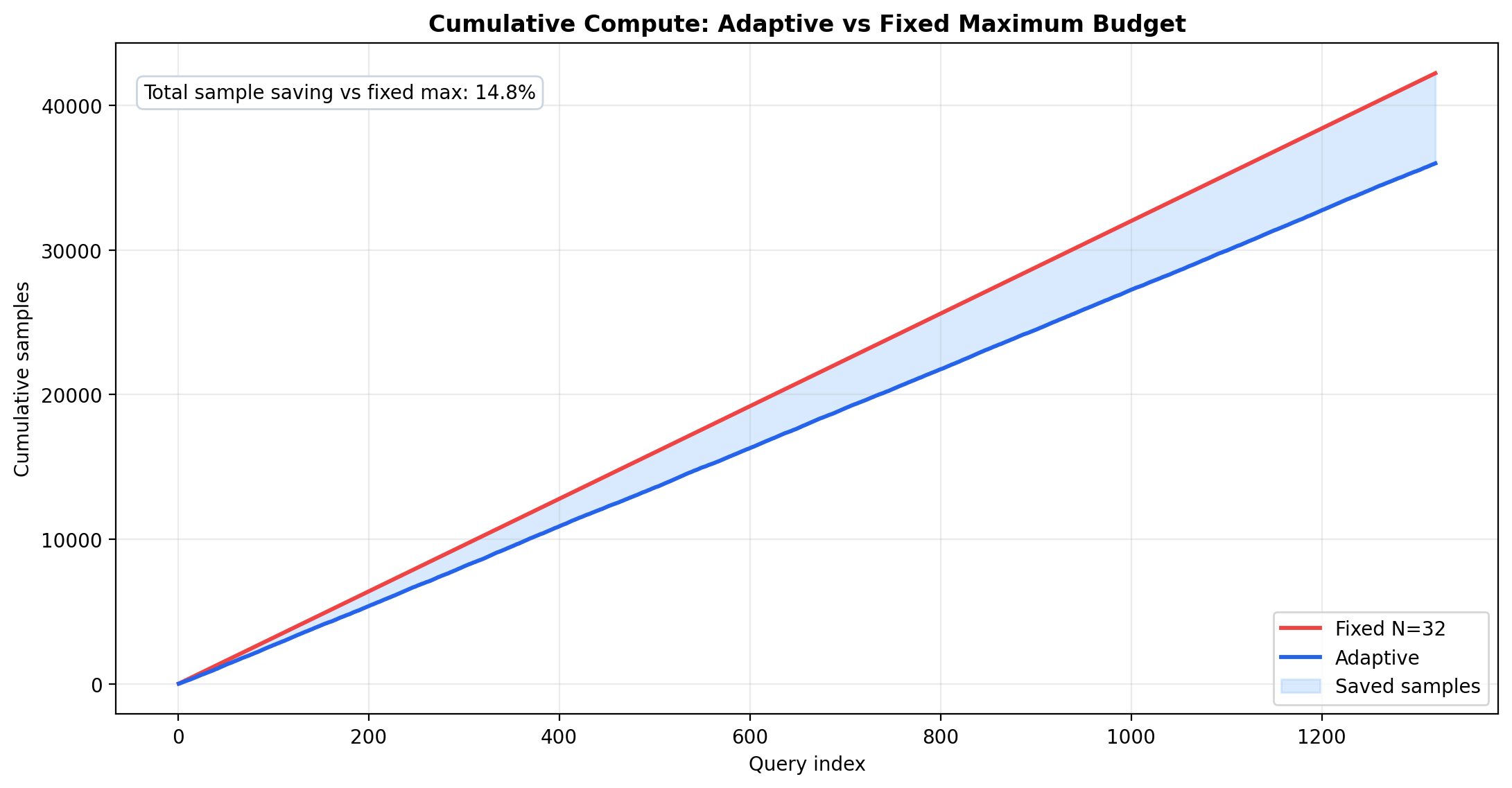}
        \caption{Qwen2.5-1.5B, \(N_{\max}=32\).}
        \label{fig:qwen-n32-compute}
    \end{subfigure}

    \caption{Cumulative compute savings for Qwen2.5-1.5B adaptive GSM8K runs. The adaptive controller saves \(14.4\%\) of samples relative to fixed \(N=16\) and \(14.8\%\) relative to fixed \(N=32\).}
    \label{fig:qwen-cumulative-compute}
\end{figure}

Figure~\ref{fig:qwen-cumulative-compute} shows smaller but still consistent savings for Qwen2.5-1.5B. Compared with Phi-3-medium, the adaptive controller stays closer to the fixed maximum budget, which means it behaves more conservatively. This is consistent with the main results: smaller or less reliable models may need more candidate samples, and the main bottleneck may shift from budget allocation to answer selection.

\section{Full Fair-Alignment Table}
\label{app:full-comparison}
Table~\ref{tab:full-comparison} reports the full fair-alignment comparison used to support the main results. This table excludes the selector-matched fixed-\(N=8\) control, which is discussed separately in the main paper, and focuses on standard fixed-budget and compute-allocation baselines. Across the reported model--dataset pairs, Adaptive improves over Best-of-\(N\), compute-optimal allocation, and self-certainty-only baselines. The main advantage is not only accuracy, but also compute allocation: Adaptive uses fewer than eight samples on average while maintaining the strongest accuracy among these baseline rows. The Adaptive rows in this table use the same fair-alignment setting as Table~\ref{tab:selector-matched}. Table~\ref{tab:selector-matched} separately reports the selector-matched fixed-\(N=8\) control, while this appendix table compares Adaptive against additional standard baselines.
\begin{table}[h]
\centering
\caption{Full fair-alignment comparison on MATH and GSM8K. Accuracy is reported with 95\% Wilson confidence intervals. Sample reduction is relative to fixed $N=8$.}
\label{tab:full-comparison}
\scriptsize
\setlength{\tabcolsep}{3pt}
\renewcommand{\arraystretch}{1.08}
\begin{tabular}{@{}lllccc@{}}
\toprule
Model & Dataset & Method & Accuracy (95\% CI) & Avg. $\bar N$ & Sample red. \\
\midrule

\multirow{12}{*}{Phi-3-mini}
& \multirow{6}{*}{MATH}
& Best-of-1 \cite{snell2024scaling}& 0.451 [0.424, 0.478] & 1.00 & 87.5\% \\
& & Best-of-3 \cite{snell2024scaling}& 0.335 [0.310, 0.361] & 3.00 & 62.5\% \\
& & Best-of-5 \cite{snell2024scaling}& 0.261 [0.238, 0.285] & 5.00 & 37.5\% \\
& & Compute-optimal \cite{snell2024scaling}& 0.415 [0.388, 0.442] & 2.64 & 67.0\% \\
& & Self-Certainty only\cite{borda} & 0.431 [0.405, 0.458] & 8.00 & 0.0\% \\
& & \textbf{Adaptive (ours)} & \textbf{0.578 [0.552, 0.605]} & \textbf{7.14} & \textbf{10.8\%} \\

\cmidrule(l){2-6}

& \multirow{6}{*}{GSM8K}
& Best-of-1 \cite{snell2024scaling}& 0.601 [0.575, 0.627] & 1.00 & 87.5\% \\
& & Best-of-3 \cite{snell2024scaling}& 0.462 [0.435, 0.489] & 3.00 & 62.5\% \\
& & Best-of-5 \cite{snell2024scaling}& 0.389 [0.363, 0.416] & 5.00 & 37.5\% \\
& & Compute-optimal \cite{snell2024scaling}& 0.503 [0.476, 0.530] & 3.00 & 62.5\% \\
& & Self-Certainty only\cite{borda} & 0.591 [0.564, 0.617] & 8.00 & 0.0\% \\
& & \textbf{Adaptive (ours)} & \textbf{0.715 [0.693, 0.742]} & \textbf{7.88} & \textbf{1.4\%} \\

\midrule

\multirow{12}{*}{Qwen2.5-1.5B}
& \multirow{6}{*}{MATH}
& Best-of-1 \cite{snell2024scaling}& 0.249 [0.226, 0.273] & 1.00 & 87.5\% \\
& & Best-of-3 \cite{snell2024scaling}& 0.193 [0.173, 0.216] & 3.00 & 62.5\% \\
& & Best-of-5 \cite{snell2024scaling}& 0.144 [0.126, 0.164] & 5.00 & 37.5\% \\
& & Compute-optimal \cite{snell2024scaling}& 0.218 [0.197, 0.241] & 2.64 & 67.0\% \\
& & Self-Certainty only \cite{borda}& 0.258 [0.235, 0.282] & 8.00 & 0.0\% \\
& & \textbf{Adaptive (ours)} & \textbf{0.293 [0.269, 0.319]} & \textbf{6.85} & \textbf{14.4\%} \\

\cmidrule(l){2-6}

& \multirow{6}{*}{GSM8K}
& Best-of-1 \cite{snell2024scaling}& 0.355 [0.329, 0.381] & 1.00 & 87.5\% \\
& & Best-of-3 \cite{snell2024scaling}& 0.265 [0.242, 0.290] & 3.00 & 62.5\% \\
& & Best-of-5 \cite{snell2024scaling}& 0.207 [0.186, 0.230] & 5.00 & 37.5\% \\
& & Compute-optimal \cite{snell2024scaling}& 0.306 [0.281, 0.331] & 3.00 & 62.5\% \\
& & Self-Certainty only\cite{borda} & 0.361 [0.335, 0.387] & 8.00 & 0.0\% \\
& & \textbf{Adaptive (ours)} & \textbf{0.464 [0.437, 0.491]} & \textbf{7.69} & \textbf{3.9\%} \\

\bottomrule
\end{tabular}
\end{table}
 The results show that fixed sampling alone is not sufficient. Best-of-3 and Best-of-5 often perform worse than Best-of-1, which suggests that adding more candidates can hurt when the selector does not aggregate them effectively. In contrast, Adaptive combines a per-prompt budget with answer-level selection, which helps it use the candidate set more reliably. The sample reduction is largest on MATH, especially for Qwen2.5-1.5B, while GSM8K receives budgets closer to the maximum. This pattern is consistent with the main paper: the controller saves more compute when more prompts can safely use smaller budgets, and behaves more conservatively when near-full aggregation is useful.

\section{Selector Sensitivity Analysis}
\label{app:selector-sensitivity}
The main results fix the answer selector to isolate the effect of adaptive budgeting. Here, we test whether the conclusion changes with different selectors. We compare self-certainty only, majority voting, and self-certainty+Borda. Fixed rows use \(N=8\); adaptive rows use the same budget policy as the main experiments.
\begin{table}[h]
\centering
\caption{Selector sensitivity analysis. Accuracy is exact-match accuracy. Adaptive sample reduction is relative to fixed \(N=8\) with the same selector.}
\label{tab:selector-sensitivity}
\small
\setlength{\tabcolsep}{4pt}
\renewcommand{\arraystretch}{1.08}
\begin{tabular}{@{}ll l l r r r@{}}
\toprule
Model & Dataset & Budget & Selector & Accuracy & Avg. $\bar N$ & Sample red. \\
\midrule
Phi-3-mini & GSM8K & Fixed \(N=8\) & Majority & 0.7248 & 8.00 & 0.0\% \\
Phi-3-mini & GSM8K & Adaptive & Majority & 0.7225 & 7.65 & 4.4\% \\
Phi-3-mini & GSM8K & Fixed \(N=8\) & Self-certainty+Borda & 0.7172 & 8.00 & 0.0\% \\
Phi-3-mini & GSM8K & Adaptive & Self-certainty+Borda & 0.7180 & 7.88 & 1.5\% \\
Phi-3-mini & GSM8K & Fixed \(N=8\) & Self-certainty only & 0.6065 & 8.00 & 0.0\% \\
\midrule
Phi-3-mini & MATH & Adaptive & Majority & 0.5853 & 6.90 & 13.7\% \\
Phi-3-mini & MATH & Fixed \(N=8\) & Self-certainty+Borda & 0.5845 & 8.00 & 0.0\% \\
Phi-3-mini & MATH & Fixed \(N=8\) & Majority & 0.5739 & 8.00 & 0.0\% \\
Phi-3-mini & MATH & Adaptive & Self-certainty+Borda & 0.5686 & 7.12 & 11.0\% \\
Phi-3-mini & MATH & Fixed \(N=8\) & Self-certainty only & 0.4496 & 8.00 & 0.0\% \\
\midrule
Qwen2.5-1.5B & MATH & Fixed \(N=8\) & Self-certainty+Borda & 0.3146 & 8.00 & 0.0\% \\
Qwen2.5-1.5B & MATH & Fixed \(N=8\) & Majority & 0.3116 & 8.00 & 0.0\% \\
Qwen2.5-1.5B & MATH & Adaptive & Self-certainty+Borda & 0.2934 & 6.85 & 14.4\% \\
Qwen2.5-1.5B & MATH & Fixed \(N=8\) & Self-certainty only & 0.2411 & 8.00 & 0.0\% \\
\bottomrule
\end{tabular}
\end{table}
Table~\ref{tab:selector-sensitivity} shows that answer selection strongly affects test-time scaling. Self-certainty alone is weaker than answer-level aggregation. Majority voting is competitive with self-certainty+Borda and is sometimes better. Adaptive majority nearly matches fixed \(N=8\) majority on GSM8K with fewer samples, and gives the best Phi-3-mini MATH result in this sweep. For Qwen2.5-1.5B on MATH, fixed \(N=8\) remains strongest. These results support the main interpretation: adaptive budgeting can reduce sampling, but the final accuracy also depends on how well the selector uses the candidate pool.
\section{Adaptive Sampling Pseudocode}
\label{app:adaptive-sampling-pseudocode}
Algorithm~\ref{alg:adaptive-sampling} summarizes the adaptive inference loop. For each prompt, the system first extracts difficulty and uncertainty signals. The adaptive policy then maps these signals to a per-prompt sample budget \(N_i\). The language model generates \(N_i\) candidate answers, and the final answer is selected by aggregating the candidates. The same trace also records the selected budget, signal values, candidate consensus, and correctness when the ground truth is available.
\begin{algorithm}[h]
\caption{Adaptive Sampling and Answer Aggregation}
\label{alg:adaptive-sampling}
\begin{algorithmic}[1]
\Require Prompt set $\mathcal{X}$, maximum sample budget $N_{\max}$, adaptive policy $\pi$, language model $M$
\Ensure Final answer $\hat{y}_i$ and trace statistics for each prompt $x_i$
\For{each prompt $x_i \in \mathcal{X}$}
    \State Compute decision signals for $x_i$:
    \Statex \hspace{\algorithmicindent} complexity $c_i$, confidence $p_i$, uncertainty $u_i$, entropy $h_i$
    \State Select the number of samples:
    \[
        N_i \gets \pi(c_i, p_i, u_i, h_i; N_{\max})
    \]
    \State Initialize candidate answer multiset $\mathcal{A}_i \gets \emptyset$
    \For{$j = 1$ to $N_i$}
        \State Generate one candidate answer:
        \[
            a_{ij} \gets M(x_i)
        \]
        \State Add $a_{ij}$ to $\mathcal{A}_i$
    \EndFor
    \State Aggregate candidates:
    \[
        \hat{y}_i \gets \operatorname{Aggregate}(\mathcal{A}_i)
    \]
    \State Record $N_i$, candidate consensus, signal values, and correctness if available.
\EndFor
\State \Return final answers and adaptive trace statistics
\end{algorithmic}
\end{algorithm}

\subsection{Example Input--Output Trace and Metric Interpretation}
\label{app:sample-input-output-metrics}

This subsection gives an illustrative example of how one prompt moves through the adaptive sampling pipeline. The example is not a new experimental result; it is a concrete trace-style explanation of the quantities reported in the appendix figures and tables.

\paragraph{Example input.}
\begin{quote}
\small
\textbf{Prompt:} A store sells 3 notebooks for \$12. If a student buys 7 notebooks at the same price per notebook, how much does the student pay?
\end{quote}

\paragraph{Adaptive decision signals.}
Before sampling multiple answers, the adaptive controller computes prompt-level signals. A simple trace for this example could look as follows:

\begin{table}[h]
\centering
\small
\caption{Illustrative adaptive decision signals for one input prompt.}
\label{tab:example-adaptive-trace}
\begin{tabular}{lrl}
\hline
Signal & Example value & Interpretation \\
\hline
Complexity $c_i$ & 0.32 & The prompt appears relatively simple. \\
Confidence $p_i$ & 0.94 & The model or controller is confident. \\
Uncertainty $u_i$ & 0.10 & The decision is low uncertainty. \\
Entropy $h_i$ & 0.41 & Candidate distribution is expected to be concentrated. \\
Selected budget $N_i$ & 1 & The controller assigns a small sample budget. \\
\hline
\end{tabular}
\end{table}

\paragraph{Example output.}
Because the prompt is estimated to be easy and high-confidence, the adaptive policy may choose only one sample. The sampled reasoning path computes the unit price as $12 / 3 = 4$ dollars per notebook and then multiplies by 7, giving $7 \times 4 = 28$. The final aggregated answer is therefore:

\begin{quote}
\small
\textbf{Model output:} The student pays \$28.\\
\textbf{Final answer:} 28
\end{quote}

For a harder prompt, the same pipeline would assign a larger $N_i$, collect multiple candidate answers, and return the answer with the strongest majority vote or consensus score. Thus, the adaptive method tries to spend few samples on easy prompts while reserving more samples for prompts with higher complexity, lower confidence, or greater uncertainty.

\paragraph{Example input.}
\begin{quote}
\small
\textbf{Prompt:} A school bought 18 boxes of pencils. Each box contains 24 pencils. The school gave 135 pencils to the fifth grade and split the remaining pencils equally among 9 classrooms. How many pencils did each classroom receive?
\end{quote}

\paragraph{Adaptive decision signals.}
Before sampling multiple answers, the adaptive controller computes prompt-level signals. A simple trace for this example could look as follows:

\begin{table}[h]
\centering
\small
\caption{Illustrative adaptive decision signals for one input prompt.}
\label{tab:example-adaptive-trace}
\begin{tabular}{lrl}
\hline
Signal & Example value & Interpretation \\
\hline
Complexity $c_i$ & 0.71 & The prompt is complex, it includes multiple arithmetic operations. \\
Confidence $p_i$ & 0.61 & The model or controller is only moderately confident. \\
Uncertainty $u_i$ & 0.37 & The decision has noticeable uncertainty. \\
Entropy $h_i$ & 0.88 & Candidate answers are expected to be more dispersed. \\
Selected budget $N_i$ & 12 & The controller assigns a larger sample budget. \\
\hline
\end{tabular}
\end{table}

\paragraph{Example output.}
Because the prompt requires multiple arithmetic steps and has only moderate confidence, the adaptive policy may choose a larger budget such as $N_i=12$. The model then generates 12 candidate reasoning paths. Some samples may make arithmetic mistakes, but the aggregation step selects the answer supported by the strongest consensus. A correct reasoning path computes the total number of pencils as $18 \times 24 = 432$, subtracts the 135 pencils given to fifth grade to obtain $432 - 135 = 297$, and then divides the remaining pencils across 9 classrooms: $297 / 9 = 33$. The final aggregated answer is therefore:

\begin{quote}
\small
\textbf{Candidate answers from 12 samples:} 33, 33, 33, 33, 33, 31, 33, 34, 33, 33, 33, 33.\\
\textbf{Model output after aggregation:} Each classroom receives 33 pencils.\\
\textbf{Final answer:} 33
\end{quote}

This example shows why a larger $N_i$ can be useful: even when a few sampled reasoning paths produce incorrect answers, the majority of samples agree on the correct value. Thus, the adaptive method tries to spend few samples on easy prompts while reserving more samples for prompts with higher complexity, lower confidence, or greater uncertainty.

\section{Discussion}
\label{sec:discussion}

The experiments highlight the importance of separating compute allocation from final answer selection. The adaptive controller decides how many candidates to generate, but the selector determines how well those candidates are used. This distinction explains why more samples do not always improve accuracy: additional candidates can help only when the selector can identify the strongest answer. If the selector is weak, extra samples can also add more plausible but wrong candidates and reduce final accuracy.

The budget distributions show that the controller behaves differently across tasks. It saves more samples on MATH, where more prompts receive budgets below \(N=8\), while it stays closer to the full budget on GSM8K. This suggests that adaptive scaling should be conservative when the task appears uncertain, rather than reducing compute aggressively for every prompt. The goal is not to minimize the number of samples at all costs, but to reduce unnecessary sampling while keeping enough candidate diversity for reliable aggregation.

The ablation results support this interpretation. Very small fixed budgets lose accuracy, which shows that candidate diversity is important. At the same time, the component ablations show that not every signal is equally useful in every setting. This means the current fuzzy controller should be viewed as an interpretable allocation policy, not as an optimized final policy. Future work can improve the selector, tune the fuzzy rules, or combine the controller with dynamic early stopping.



\small

\begin{thebibliography}{00}

\bibitem{singhc}
Singh C, Inala JP, Galley M, Caruana R, Gao J. Rethinking interpretability in the era of large language models. arXiv preprint arXiv:2402.01761. 2024.

\bibitem{bboxllm}
Chen L, Chen J, Goldstein T, Huang H, Zhou T. InstructZero: Efficient instruction optimization for black-box large language models. arXiv preprint arXiv:2306.03082. 2023.

\bibitem{wang2023selfconsistency}
Wang X, Wei J, Schuurmans D, Le Q, Chi E, Narang S, Chowdhery A, Zhou D. Self-consistency improves chain of thought reasoning in language models. ICLR, 2023.

\bibitem{snell2024scaling}
Snell C, Lee J, Xu K, Kumar A. Scaling LLM test-time compute optimally can be more effective than scaling parameters for reasoning. ICLR, 2025.

\bibitem{atomofthought}
Teng F, Shi Q, Yu Z, Zhang J, Luo Y, Wu C, Guo Z. Atom of thoughts for Markov LLM test-time scaling. arXiv preprint arXiv:2502.12018. 2025.

\bibitem{1bllm}
Liu R, Gao J, Zhao J, Zhang K, Li X, Qi B, Ouyang W, Zhou B. Can 1B LLM surpass 405B LLM? Rethinking compute-optimal test-time scaling. arXiv preprint arXiv:2502.06703. 2025.

\bibitem{mctsjudge}
Wang Y, Ji P, Yang C, Li K, Hu M, Li J, Sartoretti G. MCTS-Judge: Test-time scaling in LLM-as-a-judge for code correctness evaluation. arXiv preprint arXiv:2502.12468. 2025.

\bibitem{ttcompute}
Snell C, Lee J, Xu K, Kumar A. Scaling LLM test-time compute optimally can be more effective than scaling model parameters. arXiv preprint arXiv:2408.03314. 2024.

\bibitem{optimaltts}
Yang W, Ma S, Lin Y, Wei F. Towards thinking-optimal scaling of test-time compute for LLM reasoning. arXiv preprint arXiv:2502.18080. 2025.

\bibitem{rethinkingprompt}
Liu Y, Li Z, Fang Z, Xu N, He R, Tan T. Rethinking the role of prompting strategies in LLM test-time scaling. ACL, 2025.

\bibitem{traellm}
Gao P, Tian Z, Meng X, Wang X, Hu R, Xiao Y, Liu Y, Zhang Z, Chen J, Gao C, Lin Y. Trae Agent: An LLM-based agent for software engineering with test-time scaling. arXiv preprint arXiv:2507.23370. 2025.

\bibitem{benchmarktts}
Li X, Ming R, Setlur P, Paladugu A, Tang A, Kang H, Shao S, Jin R, Xiong C. Benchmark test-time scaling of general LLM agents. arXiv preprint arXiv:2602.18998. 2026.

\bibitem{fasttts}
Chen HM, Mo Z, Lu G, Liang S, Ma L, Luk W, Fan H. FastTTS: Accelerating test-time scaling for edge LLM reasoning. ASPLOS, 2026.

\bibitem{guitts}
Yang Y, Li D, Dai Y, Luo Z, Zhao Z, Hu Z, Huang J, Saha A, Chen Z, Xu R. GTA1: GUI test-time scaling agent. arXiv preprint arXiv:2507.05791. 2025.

\bibitem{ttsfoundationmodels}
Cong W, Zhu H, Wang P, Liu B, Xu D, Wang K, Pan DZ, Wang Y, Fan Z, Wang Z. Can test-time scaling improve world foundation model? arXiv preprint arXiv:2503.24320. 2025.

\bibitem{kinetics}
Sadhukhan R, Chen Z, Zheng H, Zhou Y, Strubell E, Chen B. Kinetics: Rethinking test-time scaling laws. arXiv preprint arXiv:2506.05333. 2025.

\bibitem{z1}
Yu Z, Wu Y, Zhao Y, Cohan A, Zhang XP. Z1: Efficient test-time scaling with code. EMNLP Industry Track, 2025.

\bibitem{selfcalib}
Huang C, Huang L, Leng J, Liu J, Huang J. Efficient test-time scaling via self-calibration. arXiv preprint arXiv:2503.00031. 2025.

\bibitem{expandperfboundaries}
Chen Z, Wang W, Cao Y, Liu Y, Gao Z, Cui E, Zhu J, Ye S, Tian H, Liu Z, Gu L. Expanding performance boundaries of open-source multimodal models with model, data, and test-time scaling. arXiv preprint arXiv:2412.05271. 2024.

\bibitem{revisito1}
Zeng Z, Cheng Q, Yin Z, Zhou Y, Qiu X. Revisiting the test-time scaling of o1-like models. ACL, 2025.

\bibitem{ttssurvey}
Zhang Q, Lyu F, Sun Z, Wang L, Zhang W, Hua W, Wu H, Guo Z, Wang Y, Muennighoff N, King I. A survey on test-time scaling in large language models. arXiv preprint arXiv:2503.24235. 2025.

\bibitem{notsay}
Turpin M, Michael J, Perez E, Bowman S. Language models don't always say what they think: Unfaithful explanations in chain-of-thought prompting. Advances in Neural Information Processing Systems. 2023 Dec 15;36:74952-65.

\bibitem{borda}
Kang Z, Zhao X, Song D. Scalable best-of-n selection for large language models via self-certainty. arXiv preprint arXiv:2502.18581. 2025 Feb 25.

\bibitem{gsm8k}
Cobbe K, Kosaraju V, Bavarian M, Chen M, Jun H, Kaiser L, Plappert M, Tworek J, Hilton J, Nakano R, Hesse C, Schulman J. Training verifiers to solve math word problems. arXiv preprint arXiv:2110.14168. 2021.

\bibitem{math}
Hendrycks D, Burns C, Kadavath S, Arora A, Basart S, Tang E, Song D, Steinhardt J. Measuring mathematical problem solving with the MATH dataset. In: Proceedings of the Neural Information Processing Systems Track on Datasets and Benchmarks. 2021.

\bibitem{sciq}
Welbl J, Liu NF, Gardner M. Crowdsourcing multiple choice science questions. arXiv preprint arXiv:1707.06209. 2017.



\end{thebibliography}
\end{document}